\documentclass[journal,web]{ieeecolor}
\usepackage{tmi}
\usepackage{cite}
\usepackage{amsmath,amssymb,amsfonts}
\usepackage{algorithmic}
\usepackage{graphicx}
\usepackage{textcomp}
\usepackage{multirow}
\usepackage{booktabs}  
\usepackage{float}
\usepackage{ulem}

\begin{document}

\title{3D Shuffle-Mixer: An Efficient Context-Aware Vision Learner of Transformer-MLP Paradigm for Dense Prediction in Medical Volume}

\author{
Jianye Pang, Cheng Jiang, Yihao Chen, Jianbo Chang, Ming Feng, Renzhi Wang and Jianhua Yao
\thanks{J. Pang is with the Department of Computer Science, Xi'an Jiaotong University, Xi'an, China (e-mail: jianye.pang97@gmail.com).}
\thanks{C. Jiang and J. Yao are with the Tencent AI Lab, Shenzhen, China (corresponding author: J. Yao, e-mail: jianhuayao@tencent.com).}
\thanks{Y. Chen, J. Chang, M. Feng and R. Wang are with the Department of Neurosurgery, Peking Union Medical College Hospital, Peking Union.}
\thanks{J. Pang and C. Jiang contributed equally to this work.}
}

\maketitle

\newcommand{\note}[1]{\textcolor{blue}{\small{\bf [ #1 --note ]}}}

\begin{abstract}

Dense prediction in medical volume provides enriched guidance for clinical
analysis. CNN backbones have met bottleneck due to lack of long-range dependencies and global context modeling power. Recent works proposed to combine vision transformer with CNN, due to its strong global capture ability and learning capability. However, most works are limited to simply applying pure transformer with several fatal flaws (i.e., lack of inductive bias, heavy computation and 
little consideration for 3D data). Therefore, designing an elegant and efficient vision transformer learner for dense prediction in medical volume is promising and challenging.
In this paper, we propose a novel 3D Shuffle-Mixer network of a new Local Vision Transformer-MLP paradigm for medical dense prediction. In our network, a local vision transformer block is utilized to shuffle and learn spatial context from full-view slices of rearranged volume, a residual axial-MLP is designed to mix and capture remaining volume context in a slice-aware manner, and a MLP view aggregator is employed to project the learned full-view
rich context to the volume feature in a view-aware manner. Moreover, an Adaptive Scaled Enhanced Shortcut is proposed for local vision transformer to enhance feature along spatial and channel dimensions adaptively, and a CrossMerge is proposed to skip-connects the multi-scale feature appropriately in the pyramid architecture.
Extensive experiments demonstrate the proposed model outperforms other state-of-the-art medical dense prediction methods.
\end{abstract}  

\begin{IEEEkeywords}
Dense prediction in medical volume, context-aware, window-based multi-head self-attention, Local Vision Transformer-MLP, Adaptive Scaled Shortcut.
\end{IEEEkeywords}

\section{Introduction}
\label{sec:introduction}

\IEEEPARstart{D}{ense} prediction 
plays an important role in medical image analysis since it
connects with a variety of practical clinical applications~\cite{wang2018deepigeos,wang2021review,huang2021dense,wang2019deeply} throughout the entire medical imaging community. Compared with classification or recognition, dense prediction
(e.g. segmentation, registration, 
synthesis, reconstruction, etc)
aims to extract contextual information and perform pixel-level classification or regression on the whole image relying on high-resolution and multi-scale features, which calls for higher requirements of model design. Besides, 3D medical volumetric data from like Computed tomography (CT) and Magnetic resonance imaging (MRI) are harder to deal with and far beyond 2D images. Accurate dense prediction in medical volume
provides enriched guidance as a fundamental enabler
for automated clinical diagnosis and analysis. Thus, many works~\cite{balakrishnan2019voxelmorph,wang2019volumetric,ebner2018automated} have focused on various dense prediction tasks in medical volume.
Although the overall frameworks of most tasks are different, the backbones used in them are 3D U-Net~\cite{cciccek20163d} and its variants~\cite{zhou2018unet++} which commonly adopt the pyramid structure based on CNN.

Since the volumetric data has one more dimension than 2D image, it contains richer semantic context in the overall volume. Besides, MRI or CT sometimes comes in multiple modalities with complex composition form. Fully learning context is necessary for dense prediction and raises great challenges to pure CNN models where CNN is not context-aware and lacks the ability to model long distance dependencies on global context. 
Despite 3D CNN models have achieved great success in medical volume, due to the fact that the amount of 3D volume is much less than 2D scan, 3D CNN models sometimes perform poorly on small datasets, even worse than 2D CNN. Moreover, the computational complexity of 3D CNN is much higher, which leads to limited freedom in structural design.
It is noteworthy that the number of applications~\cite{dosovitskiy2020image,d2021convit,tolstikhin2021mlp} on vision transformer or vision MLP in general vision tasks have been exploded very recently. 
To get rid of the above dilemma of pure CNN for dense prediction, how to use these remarkable architectures to break the bottleneck of CNN backbones has became a hot topic in medical image analysis.

Taking advantage of strong learning capability of vision transformer, a large portion of medical dense prediction~\cite{chen2021transunet,valanarasu2021medical,chen2021vit,wang2021ted} works on adding pure transformer to pyramid CNN recently.
However, due to the fatal flaws of pure transformer from large dataset requirement, lack of inductive bias, heavy computation and over-smoothing degradation, pure vision transformer is underwhelming for dense prediction.
CeiT~\cite{yuan2021incorporating} 
proposed using depth-wise convolution into transformer block to fuse information from different patches and increase locality. DeepViT~\cite{zhou2021deepvit} and Refiner~\cite{zhou2021refiner} proposed re-attention linear transformations to add diversity and alleviate degradation. One of the most notable improvement is the local vision transformer~\cite{han2021demystifying} which proposed window-based multi-head self-attention (W-MSA) with less computation burden and more inductive bias. It dramatically improved the performance on dense prediction and applicability to small datasets which further taps transformer's potentials. However, currently there are very few studies~\cite{cao2021swin,wu2021hepatic} considering local vision transformer, at best, but are only simple to introduce existing Swin module~\cite{liu2021swin} into the pyramid structure. Furthermore, most works of local vision transformer are only proposed for 2D image. How to elegantly design an efficient vision learner specially suitable for 3D medical volume leveraging local vision transformer to solve the dense prediction task is challenging which hasn't been considered and addressed.

To this end, we propose a novel 3D Shuffle-Mixer of a new Local Vision Transformer-MLP paradigm for dense prediction in medical volume which efficiently learns volume context via adequate information shuffle and mixing. Concretely, we first rearrange volume into different view slices and
learn the rich spatial context by W-MSA 
with transpose shuffle 
to build cross-window spatial connections. Then the remaining context along third dimension is captured by a slice-aware MLP mixing, which is proposed with an residual axial-MLP and absolute position embedding (APE) on each slice feature to enable the ability to distinguish between slice context. 
Finally volume context from all views is projected to the complete volume feature through a view-aware aggregator with APE for each view to distinguish between view context. 
For one 3D Shuffle-Mixer block, both W-MSA and MLP share parameters across views.
We also propose an Adaptive Scaled Enhanced Shortcut (ASES) for window-based local vision transformer to adaptively rectificate and refine decoupled features on spatial and channel dimensions. Furthermore, a CrossMerge is proposed to skip-connect and merge view context from encoder to decoder elegantly. The effectiveness of our method is validated by comparing it with various existing state-of-the-art (SOTA) methods on three medical dense prediction tasks.

The main contributions of this paper are summarized as follows.
\begin{itemize}
    \item Different from most existing CNN-Pure Transformer hybrid approaches, we present a novel context-aware vision learner of a new Local Vision Transformer-MLP paradigm for dense prediction in medical volume.
    \item We present an ASES shortcut for window-based local vision transformer to capture more precise and augmented features enhancing along spatial and channel dimensions adaptively, as well as a CrossMerge skip-connection suitable for pyramid transformer model.
    \item We conduct extensive validations on one private shape prediction dataset and two pubilc popular segmentation datasets. Our model achieves a great performance gain, surpassing the SOTA methods by a large margin.
\end{itemize}

The rest of this paper is organized as follows. We introduce the related works in Section~\ref{sec:related works}. Then, we elaborate the framework of our 3D Shuffle-Mixer with ASES shortcut and CrossMerge skip-connection in Section~\ref{sec:method}.
We further present the analysis and discussions of experimental results 
in Section~\ref{sec:experiments}.
Finally, we conclude the paper in Section~\ref{sec:conclusion}.

\section{Related Works}
\label{sec:related works}
In this section, we first discuss some local vision transformer architectures in general vision tasks and then take the 3D segmentation as an example to review some related works of the recent popular CNN-Transformer (CNN-TR) hybrid backbones for 3D medical image analysis since they're commonly used architectures among various medical downstream dense tasks.

\textbf{Local Vision Transformers in General Vision Tasks: }
The first work to show vanilla transformer can compete with SOTA CNNs on large datasets in general vision tasks is ViT~\cite{dosovitskiy2020image} which
has become the standard benchmark for vision transformer. 
Local vision transformers improve a lot over pure transformers. They reduce complexity to linear by computing self-attention only within each local window. Moreover, various methods of cross-window connections make up for prior knowledge of locality and translation invariance which contributes to the success of local vision transformers in dense prediction tasks. 

Swin Transformer~\cite{liu2021swin} (Swin-TR) first proposed shifted windows to bridge window information to encourage inductive bias, and introduced relative position bias into window-based local vision transformer. Twins~\cite{chu2021twins} combined local attention and global attention alternately in each stage to enhance modeling power. Shuffle Transformer~\cite{huang2021shuffle} (Shuffle-TR) shuffled fixed windows to make information flow across windows. PVTv1~\cite{wang2021pyramid} proposed using pyramid structure in transformer and PVTv2~\cite{wang2021pvtv2} proposed overlapping patch embedding to enlarge patch window. MSG-Transformer~\cite{fang2021msg} shuffled the message token added on each window for message passing across windows. The most recent SOTA in general vision tasks is CSWin Transformer~\cite{dong2021cswin} (CSWin-TR) which proposed cross-shaped window self-attention to enlarge receptive field. 
Besides, there're several works~\cite{tolstikhin2021mlp,lian2021mlp,chen2021cyclemlp} on vision MLP which used concise structures, but trained on larger datasets to achieve the similar performance.
However, above models are all designed for 2D natural images and there is few research for 3D input, e.g. Video Swin Transformer~\cite{liu2021video} (Video Swin-TR) which extends to 3D shifted windows directly. 

\textbf{CNN-Transformer Backbones in 3D Medical Image Analysis: }
3D segmentation is one of the most general and representative tasks in medical image analysis. 
Well-performed backbones for 3D medical image analysis consist of several newly proposed CNN-TR hybrid models.

Segtran~\cite{li2021medical} introduced squeeze-and-expansion into transformer and a new learnable sinusoidal position encoding for imposing continuity inductive bias. U-Transformer~\cite{petit2021u} utilized self- and cross-attention into U-Net. MedT~\cite{valanarasu2021medical} proposed using gated axial attention in transformer. TransUNet~\cite{chen2021transunet}, CoTr~\cite{xie2021cotr} and TransBTS~\cite{wang2021transbts} replaced CNN encoder in U-Net with hybrid CNN-TR which connected to pure transformer at the end of CNN to make a strong encoder. UNETR~\cite{hatamizadeh2021unetr} directly replaced whole CNN encoder with cascaded pure transformer encoder to capture multi-scale features. Swin-Unet~\cite{cao2021swin} simply brought Swin block from Swin-TR to build a U-shaped encoder-decoder symmetric architecture to improve feature representation. 
Apart from the segmentation, there're also similar CNN-TR backbones~\cite{tao2021spine,luo20213d} in 3D medical detection, reconstruction, etc.
However, above models simply used pure transformer or directly copied existing module in the pyramid implementation. Neither proposed a specially designed local vision transformer block for medical images. Besides, only a few models took 3D volume into account, e.g. Segtran, TransBTS and UNETR, while most models are simply based on 2D scan processing and therefore generally perform less well.

\section{Methodology}
\label{sec:method}
\subsection{Overview of the Proposed Method}
\begin{figure}[H]
    \centering
    \includegraphics[width=0.5\textwidth, trim={520 170 610 120}, clip]{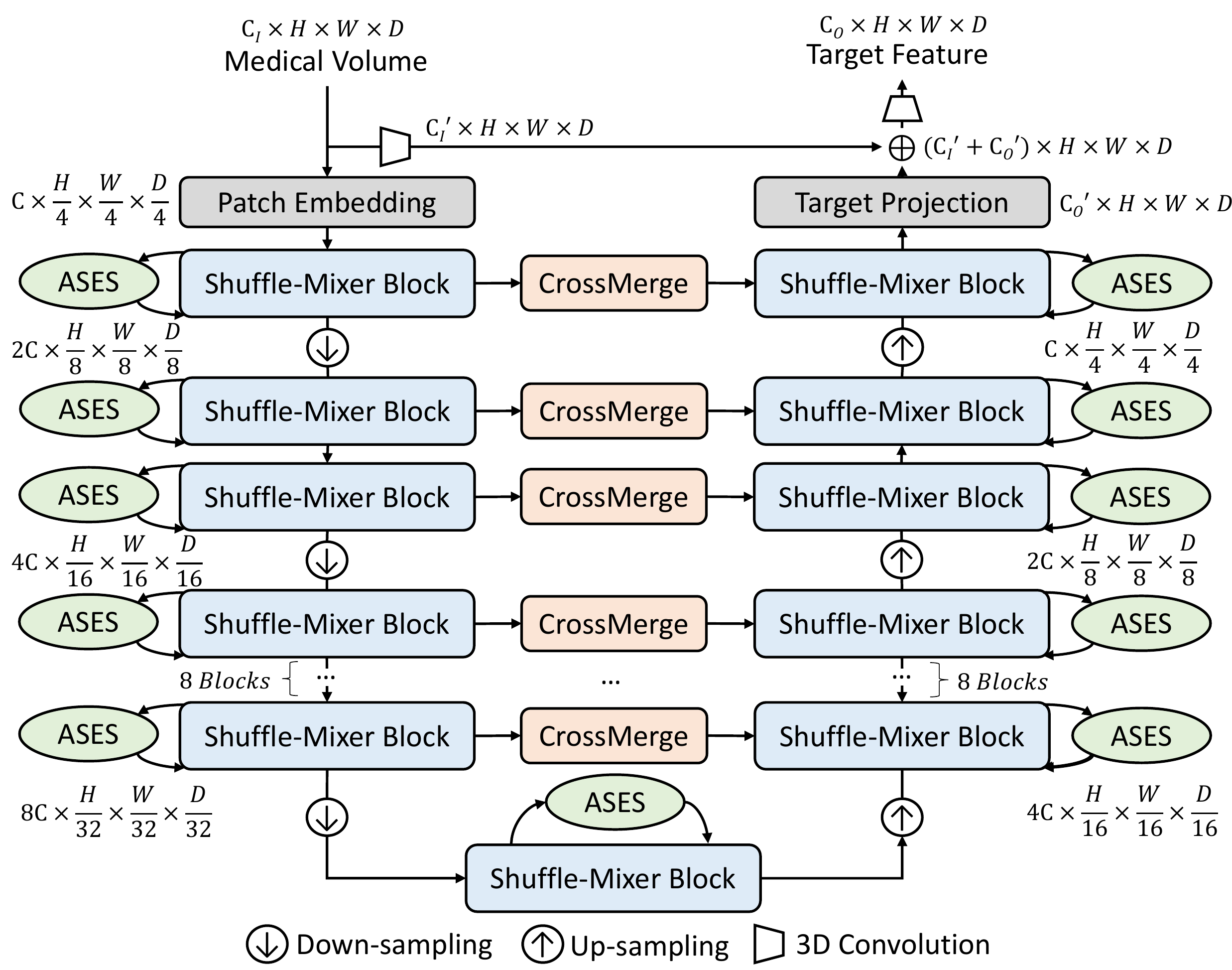}
    \caption{Overview of the proposed method. The channel of input volume is $C_{i}$ and output target is $C_{o}$.}
    \label{fig:arch-overview}
\end{figure}

\begin{figure*}
    \centering
    \includegraphics[width=1.0\textwidth, trim={190 20 190 20}, clip]{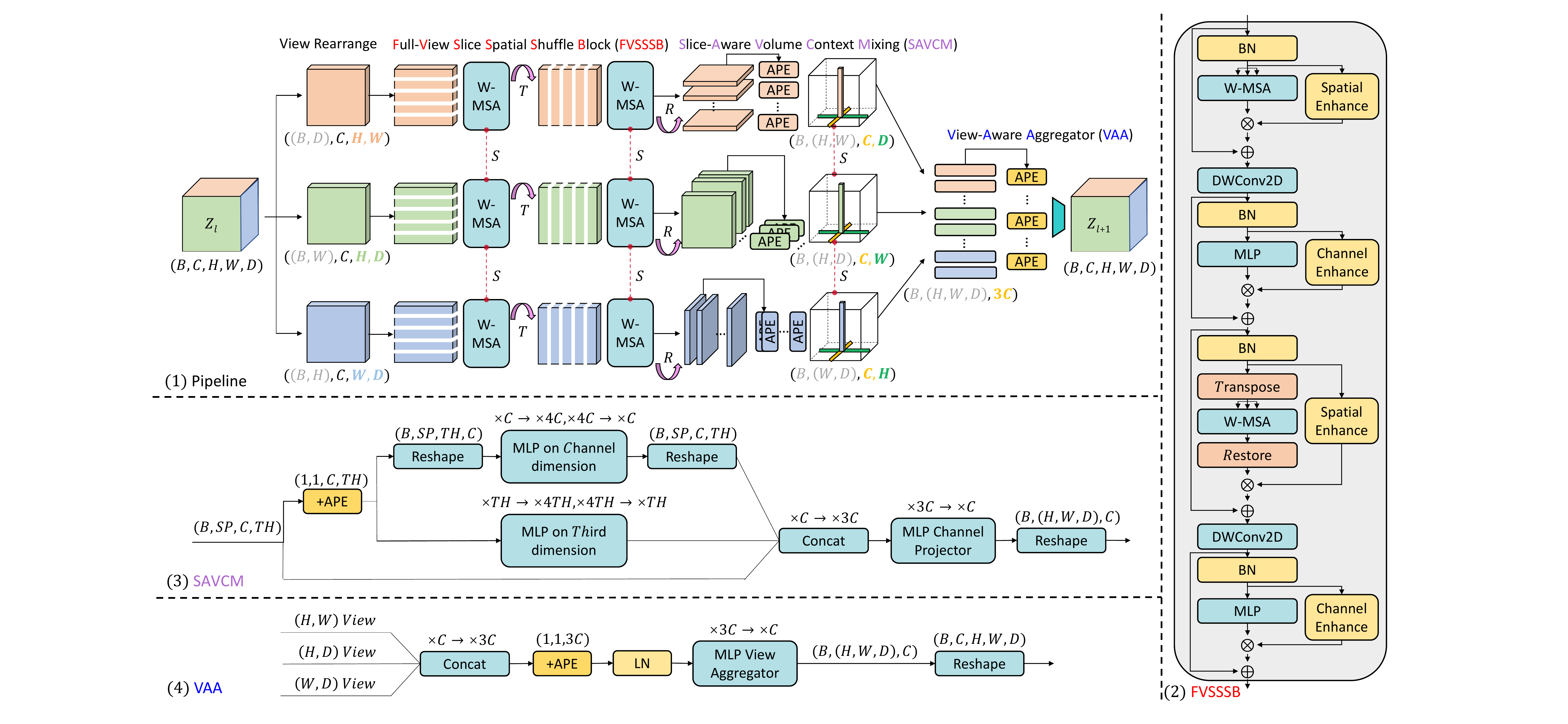}
    \caption{Architecture of 3D Shuffle-Mixer block. Overall Pipeline of 3D Shuffle-Mixer block is shown in (1), which consists of three sequential parts: Details of Full-View Slice Spatial Shuffle Block with ASES are shown in (2), where ASES is represented as Spatial Enhance and Channel Enhance which will be elaborated in Section~\ref{sec:ASES}; Details of Slice-Aware Volume Context Mixing are shown in (3); Details of View-Aware Aggregator are shown in (4). *In (1), arrow $T$ denotes transpose shuffle operation, arrow $R$ denotes rotation restore operation and red dotted line denotes parameter sharing between connected targets. In (3), $SP$ denotes 2D spatial dimension, e.g. $(H,W)$ and $TH$ denotes third dimension, e.g. $D$.}
    \label{fig:arch-shufflemixer}
\end{figure*}

Fig.~\ref{fig:arch-overview} shows the overall framework of the proposed method which is a pyramid architecture. Our model consists of 3D Shuffle-Mixer blocks (Fig.~\ref{fig:arch-shufflemixer}) with ASES (Fig.~\ref{fig:arch-ases}),
and CrossMerge (Fig.~\ref{fig:arch-crossmerge}) to skip-connect and merge multi-scale context between each encoder and decoder block. We adopt convolutional stem~\cite{xiao2021early}  as patch embedding for non-overlapping token embedding and training stability. We utilize 3D convolution with kernel size 3 in down-sampling, transposed convolution with kernel size 2 in up-sampling and kernel size 1 in target projection to facilitate information flow between non-overlapping windows.

\subsection{3D Shuffle-Mixer}

\subsubsection{Full-View Slice Spatial Shuffle Block}
We utilize W-MSA to obtain global and local spatial contextual information first. The input feature $\mathbf{Z}$ is partitioned into non-overlapping window feature $\mathbf{Z}^{i}$. $\mathbf{W}_{k}^{Q}$, $\mathbf{W}_{k}^{K}$ and $\mathbf{W}_{k}^{V}$ are query, key and value matrices for $k$-th head. In details, computing the $k$-th head self-attention $\mathbf{H}_{k}$ and output $\mathbf{W\text{-}MSA}(\mathbf{Z})$ can be defined as:
\begin{equation}
\begin{aligned}
&\mathbf{Z}=\left\{\mathbf{Z}^{1}, \mathbf{Z}^{2}, \ldots, \mathbf{Z}^{N}\right\}, N=H W / M^{2} \\
&\mathbf{H}_{k}^{i}=\mathbf{Attention}(\mathbf{Z}^{i} \mathbf{W}_{k}^{Q}, \mathbf{Z}^{i} \mathbf{W}_{k}^{K}, \mathbf{Z}^{i} \mathbf{W}_{k}^{V}), \quad i=1, \ldots, N \\
&\mathbf{H}_{k}=\left\{\mathbf{H}_{k}^{1}, \mathbf{H}_{k}^{2}, \ldots, \mathbf{H}_{k}^{N}\right\} \\
&\mathbf{W\text{-}MSA}(\mathbf{Z})=\mathbf{Concat}[\mathbf{H}_{1}, \mathbf{H}_{2}, \ldots, \mathbf{H}_{k}] \mathbf{W}^{H} 
\end{aligned}
\end{equation}
Where $M$ denotes window size. All heads are concatenated and linearly projected to the output with relative position encoding~\cite{liu2021swin} (RPE) 
to be window-aware of spatial positions. The attention is defined as:
\begin{equation}
\mathbf{ Attention }(\mathbf{Q}, \mathbf{K}, \mathbf{V})=\mathbf{SoftMax}(\mathbf{Q K}^{T} / \sqrt{d_{k}}+\mathbf{RPE}) \mathbf{V}
\end{equation}
However, pure W-MSA lacks window connections, hence
we utilize 2D spatial transpose shuffle operation inspired by~\cite{huang2021shuffle,fang2021msg,tolstikhin2021mlp} into W-MSA to fully build long-range cross-window connections. 
Besides, it can only handle 2D spatial well, which will completely ignore the context of the third dimension like depth when applied to 3D volume. 
Therefore, we propose a full-view slice spatial shuffle 
block to rearrange and extract redundant information residing in full-view spatial slices without complex and time-consuming operation like pure 3D convolution.

As shown in Fig.~\ref{fig:arch-shufflemixer}-(1),(2), we consider slices from each two-dimensional direction as one view of the 3D volume. Thus, a 3D volume $(H, W, D)$ has three kinds of views $(H, W)$, $(H, D)$ and $(W, D)$ which can be directly acquired by rearrange operation. 
Therefore, we have all slices from 3D volume to learn pure W-MSA and employ transpose shuffle $T$ with rotation restore $R$
on slices in each view to learn shuffle W-MSA respectively. 
W-MSA learns normalized features through batch norm (BN) first, and a 2D depth-wise convolution (DWConv2D) with 5$\times$5 kernel and 1 stride strengthens spatial neighbor locality on the slice, followed by a multi-layer perceptron (MLP) which maps features to high dimension and back for refining.
The computation can be represented as:
\begin{equation}
\label{fomula:(1)}
\begin{aligned}
&\mathbf{X}_{l}^{v}=\mathbf{W\text{-}MSA}\left(\mathbf{B} \mathbf{N}\left(\mathbf{Z}_{l}^{v}\right)\right)+\mathbf{Z}_{l}^{v} \\
&\mathbf{Y}_{l}^{v}=\mathbf{DWConv2D}\left(\mathbf{X}_{l}^{v}\right)+\mathbf{X}_{l}^{v} \\
&{\mathbf{Z}_{l}^{v}}^{'}=\mathbf{M L P}\left(\mathbf{B} \mathbf{N}\left(\mathbf{Y}_{l}^{v}\right)\right)+\mathbf{Y}_{l}^{v} \\
&{\mathbf{X}_{l}^{v}}^{'}=\mathbf{R}(\mathbf{W\text{-}M S A}(\mathbf{T}(\mathbf{B N}({\mathbf{Z}_{l}^{v}}^{'}))))+{\mathbf{Z}_{l}^{v}}^{'} \\
&{\mathbf{Y}_{l}^{v}}^{'}=\mathbf{DWConv2D}({\mathbf{X}_{l}^{v}}^{'})+{\mathbf{X}_{l}^{v}}^{'}  \\
&{\hat{\mathbf{Z}}_{l}^{v}}=\mathbf{M L P}(\mathbf{B} \mathbf{N}({\mathbf{Y}_{l}^{v}}^{'}))+{\mathbf{Y}_{l}^{v}}^{'} \\
&v=\{ 1, 2, 3 \}
\end{aligned}
\end{equation}
Where $\mathbf{Z}^{v}_{l}$ denotes the input volume feature comes from view $v$ in $l$-th block for W-MSA to learn.
Noticed that the parameters of each W-MSA are shared among full-view in each block. On one hand, the computation is drastically reduced. On the other hand, scalability and generalization increase since more inductive bias are brought into vision transformer. 

\subsubsection{Slice-Aware Volume Context Mixing}
Full-view information of the 3D volume has been obtained as the output of slice spatial shuffle block. However, in each single view, the output only contains spatial information of 2D slices. To take into account the context among slices, A.K.A along the third dimension, we propose a slice-aware volume context mixing to capture remaining context.

As shown in Fig.~\ref{fig:arch-shufflemixer}-(3), the slice-aware volume context mixing contains two branches. In one branch, we add absolute position embedding~\cite{ke2020rethinking} (APE) which is an automatic learnable parameter on each slice feature to inject positional information and help distinguish slices.
Then the slice-aware context is processed by an axial-MLP:
\begin{equation}
\begin{aligned}
&\mathbf{A}_{l+1}^{v}=\hat{\mathbf{Z}}_{l+1}^{v}+\mathbf{APE}_{s} \\
&\mathbf{Axial\text{-}MLP}(\hat{\mathbf{Z}}_{l+1}^{v})=[\mathbf{MLP}_{st}({\mathbf{A}_{l+1}^{v}}); \mathbf{MLP}_{sc}({\mathbf{A}_{l+1}^{v}})] \\
&\hat{\mathbf{A}}_{l+1}^{v}=\mathbf{MLP}_{cp}(\mathbf{Concat}[\mathbf{Axial\text{-}MLP}(\hat{\mathbf{Z}}_{l+1}^{v}); \hat{\mathbf{Z}}_{l+1}^{v}]) \\
\end{aligned}
\end{equation}
Where $s$ denotes slice-aware, subscript $st$ and $sc$ denote slice-aware along third dimension and channel, and $cp$ denotes channel projector.
The axial-MLP is used to linearly map on the third dimension and channel in parallel. In another branch, we use residual shortcut to maintain original feature as identity mapping. The parameters of axial-MLP are shared between among the three views as well.

In contrast to only simple spatial feature~\cite{chen2021transunet,cao2021swin} learned based on 2D scan, we capture the 3D context with concise computational operations. In addition, 
we introduce position information between slices to enable the model to be slice aware.

\subsubsection{View-Aware Aggregator}
Rich 3D volume context has been captured 
in each single view through slice-aware volume context mixing. To aggregate the full-view together and obtain the final volume feature of 3D Shuffle-Mixer at $l\text{+}1$-th block, we propose a view-aware aggregator to project features of all views 
along channel dimension 
in the end of the 3D Shuffle-Mixer block.

As shown in Fig.~\ref{fig:arch-shufflemixer}-(4), we first concatenate full-view input along channel dimension and adds APE on them, followed by a layer norm (LN) to normalize view-aware features. 
Then a MLP view aggregator restore the feature to original channel dimension. The computation is represented as:
\begin{equation}
\begin{aligned}
&\mathbf{V}_{l+1}^{v}=\hat{\mathbf{A}}_{l+1}^{v}+\mathbf{APE}_{vi} \\
&\mathbf{Z}_{l+1}=\mathbf{MLP}_{va}(\mathbf{LN}(\mathbf{Concat}[\mathbf{V}_{l+1}^{1}; \mathbf{V}_{l+1}^{2}; \mathbf{V}_{l+1}^{3}])) \\
\end{aligned}
\end{equation}
Where $vi$ denotes view-aware and subscript $va$ denotes view aggregator.
With addition of APE, MLP view aggregator is aware of the specific view.

\subsection{Adaptive Scaled Enhanced Shortcut for Window-based Local Vision Transformer}
\label{sec:ASES}
\begin{figure}
    \centering
    \includegraphics[width=0.5\textwidth, trim={130 20 140 20}, clip]{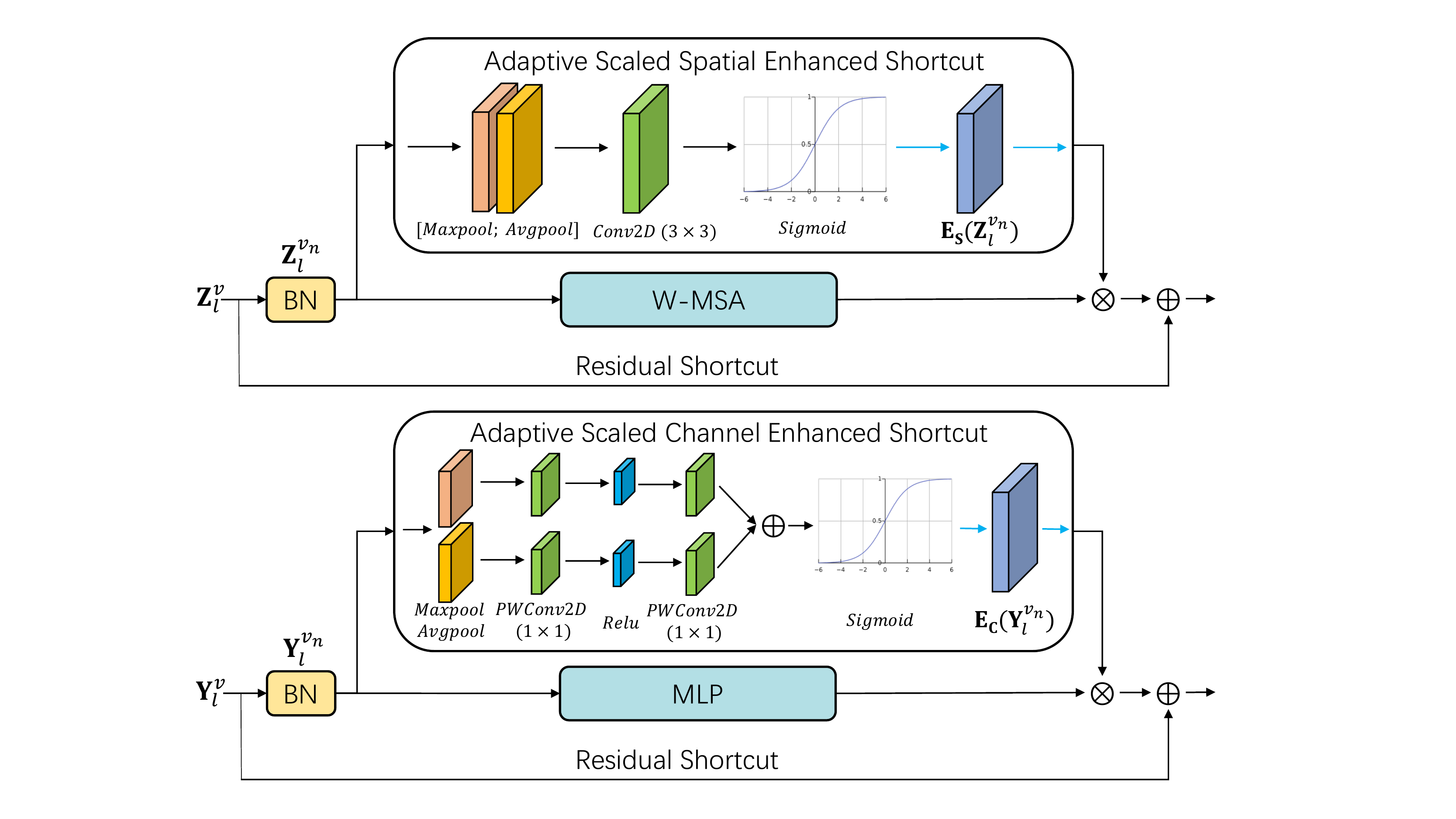}
    \caption{Architecture of Adaptive Scaled Enhanced Shortcut for Window-based Local Vision Transformer, which consists of spatial enhance and channel enhance. \color{cyan}{Cyan} \color{black}{arrow represents ASES shortcut's output}.}
    \label{fig:arch-ases}
\end{figure} 
To further increase learning capacity 
of local vision transformer, we propose an ASES for window-based local vision transformer which can enhance feature representation adaptively using its original input meanwhile only costs a little extra computation. ASES is an improvement based on Formula~(\ref{fomula:(1)}), where
the computation is represented as:
\begin{equation}
\begin{aligned}
&\mathbf{Z}_{l}^{v_{n}}=\mathbf{B N}\left(\mathbf{Z}_{l}^{v}\right) ; \mathbf{Y}_{l}^{v_{n}} =\mathbf{B N}\left(\mathbf{Y}_{l}^{v}\right) \\
&\mathbf{W\text{-}MSA^{E}}(\mathbf{Z}_{l}^{v_{n}})=\mathbf{E}_{\mathbf{s}}(\mathbf{Z}_{l}^{v_{n}}) \otimes \mathbf{W\text{-}MSA}(\mathbf{Z}_{l}^{v_{n}}) + \mathbf{Z}_{l}^{v} \\
&\mathbf{MLP^{E}}(\mathbf{Y}_{l}^{v_{n}})=\mathbf{E}_{\mathbf{c}}(\mathbf{Y}_{l}^{v_{n}}) \otimes \mathbf{MLP}(\mathbf{Y}_{l}^{v_{n}}) + \mathbf{Y}_{l}^{v} \\
\end{aligned}
\end{equation}
Where $\mathbf{Z}_{l}$ and $\mathbf{Y}_{l}$ are the original input features for the W-MSA and MLP in the $l$-th block respectively. $\otimes$ denotes element-wise multiplication. 
$\mathbf{E}_{\mathbf{s}}$ and $\mathbf{E}_{\mathbf{c}}$ are spatial enhanced shortcut and channel enhanced shortcut, which are detailed shown in Fig.~\ref{fig:arch-ases} (also in Fig.~\ref{fig:arch-shufflemixer}-(2)) and computed as:
\begin{equation}
\begin{aligned}
\mathbf{E}_{\mathbf{s}}(\mathbf{Z}_{l}^{v_{n}}) &=\sigma\left(\mathbf{Conv2D}([\mathbf{AvgPool}(\mathbf{Z}_{l}^{v_{n}}) ; \mathbf{Max} \mathbf{Pool}(\mathbf{Z}_{l}^{v_{n}}])\right) \\
&=\sigma\left(\mathbf{f^{3 \times 3}}\left(\mathbf{Concat}\left[{(\mathbf{Z}_{l}^{v_{n}})}^{\mathrm{avg}} ; {(\mathbf{Z}_{l}^{v_{n}})}^{\mathrm{\max}}\right]\right)\right) \\
\end{aligned}
\end{equation}
\begin{equation}
\begin{aligned}
\mathbf{E}_{\mathbf{c}}(\mathbf{Y}_{l}^{v_{n}}) &=\sigma(\mathbf{PWConv2D}(\mathbf{AvgPool}(\mathbf{Y}_{l}^{v_{n}})) + \\
&\mathbf{PWConv2D}(\mathbf{MaxPool}(\mathbf{Y}_{l}^{v_{n}}))) \\
&=\sigma(\mathbf{f^{1 \times 1}}((\mathbf{Y}_{l}^{v_{n}})^{\mathrm{avg}})+\mathbf{f^{1 \times 1}}((\mathbf{Y}_{l}^{v_{n}})^{\mathrm{\max}})) \\ 
\end{aligned}
\end{equation}
Where PWConv2D denotes point-wise convolution with $1 \times 1$ kernel. In the window-based local vision transformer equipped with ASES, spatial enhancing on W-MSA learns an adaptive scaled factor $\mathbf{E}_{\mathbf{s}}(\mathbf{Z}_{l})$ from input and multiply to the feature learned by W-MSA to rectificate along spatial.
Channel enhancing on MLP gives each channel with various intensity $\mathbf{E}_{\mathbf{c}}(\mathbf{Y}_{l})$ 
computed from input
and multiply to the feature learned by MLP to rectificate along channel. 
Therefore, two shortcut enhancements 
rectificate and augment the visual features using its own original information without bringing extra complicated feature transformations.  

The network compositions are inspired by CBAM~\cite{woo2018cbam} which served as a simple kind of attention implicitly, while it serves for a quite different motivation in our ASES here, which is naturally adapt to window-based local vision transformer to explicitly and adaptively enhance the decoupled features. 
Besides, compared to recent works~\cite{zhou2021deepvit,zhou2021refiner,tang2021augmented} 
using additional branch of designed transformation to diverse transformer features as augmented measure,
they're usually hard to determine what kind of transformation or
how many extra branches  
are needed
and undoubtedly require lots of extra computation. While ASES greatly alleviate these problems.

\subsection{CrossMerge Skip-Connection}
\begin{figure}
    \centering
    \includegraphics[width=0.5\textwidth, trim={80 50 140 20}, clip]{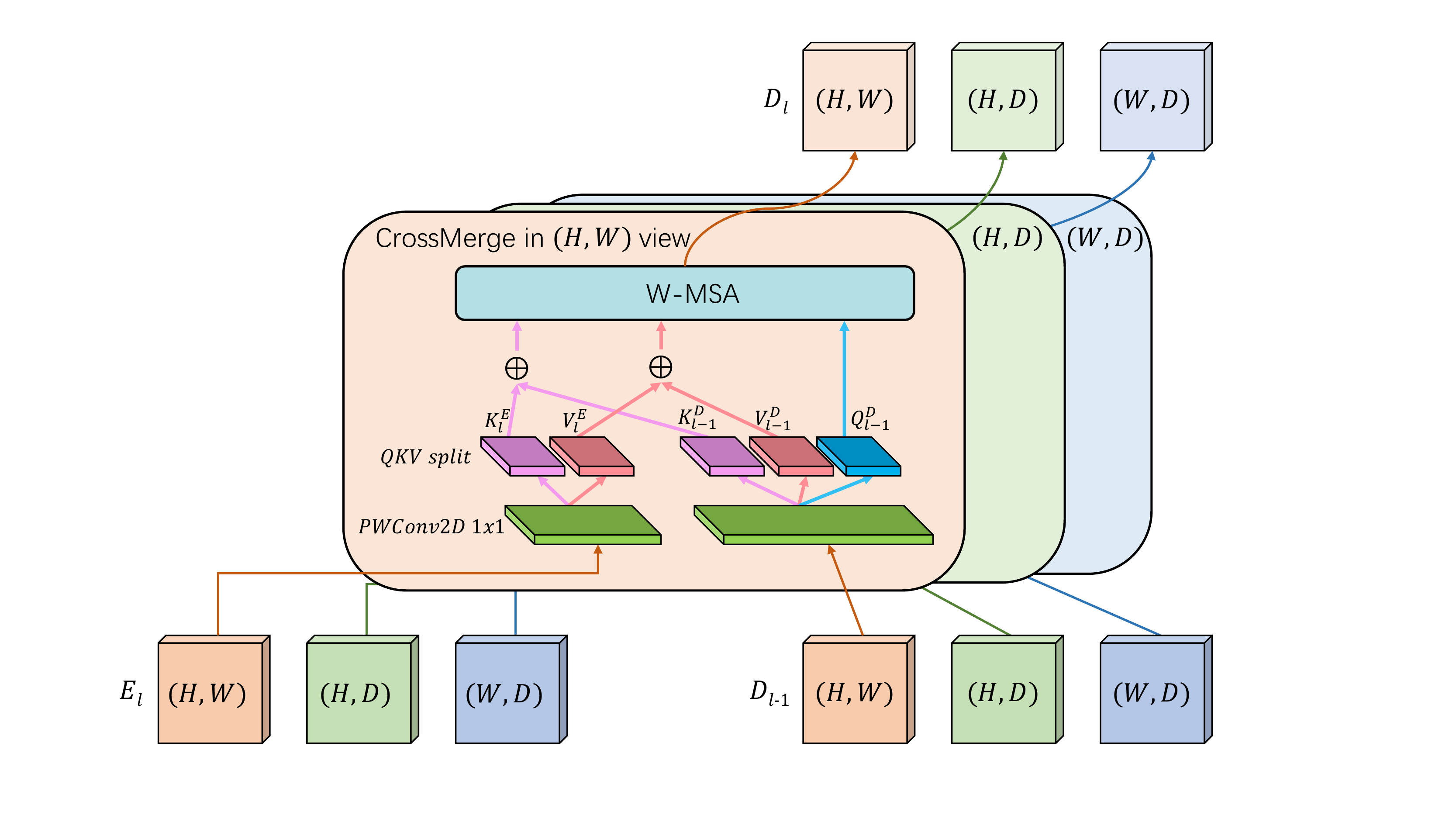}
    \caption{Architecture of CrossMerge Skip-Connection, which is interweaved across from encoder to decoder for merging information individually in each view.}
    \label{fig:arch-crossmerge}
\end{figure} 
To investigate a more appropriate skip-connection form applicable for pyramid vision transformer architecture to transfer multi-scale features from the encoder to the decoder in each block, 
meanwhile to better cope with 
merging between query, key and value, we design a new skip-connection form named as CrossMerge shown in Fig.~\ref{fig:arch-crossmerge}, which is computed as:
\begin{equation}
\begin{aligned}
&\langle \mathbf{K}^{\mathbf{E}^{v}}_{l}, \mathbf{V}^{\mathbf{E}^{v}}_{l}\rangle=\mathbf{Split}(\mathbf{PWConv2D}(\mathbf{E}_{l}^{v})) \\
&\langle \mathbf{Q}^{\mathbf{D}^{v}}_{l-1}, \mathbf{K}^{\mathbf{D}^{v}}_{l-1}, \mathbf{V}^{\mathbf{D}^{v}}_{l-1}\rangle=\mathbf{Split}(\mathbf{PWConv2D}(\mathbf{D}_{l-1}^{v})) \\
&{\mathbf{K}^{\mathbf{D}^{v}}_{l-1}}^{'}=\mathbf{K}^{\mathbf{E}^{v}}_{l}+\mathbf{K}^{\mathbf{D}^{v}}_{l-1}; {\mathbf{V}^{\mathbf{D}^{v}}_{l-1}}^{'}=\mathbf{V}^{\mathbf{E}^{v}}_{l}+\mathbf{V}^{\mathbf{D}^{v}}_{l-1}\\
&\mathbf{ CrossMerge }(\mathbf{E}_{l}^{v},\mathbf{D}_{l-1}^{v})=\mathbf{ W\text{-}MSA }({\mathbf{Q}^{\mathbf{D}^{v}}_{l-1}}, {\mathbf{K}^{\mathbf{D}^{v}}_{l-1}}^{'}, {\mathbf{V}^{\mathbf{D}^{v}}_{l-1}}^{'})
\end{aligned}
\end{equation}
Where $\mathbf{E}_{l}^{v}$ and $\mathbf{D}_{l-1}^{v}$ denotes feature from view $v$ in $l$-th block of encoder and view $v$ in $l\text{-}1$-th block of decoder.
We use PWConv2D to multiply feature channels for stability and split as $\langle \mathbf{Q,K,V}\rangle$ respectively
in pyramid structure.
$\mathbf{K}$ and $\mathbf{V}$
from both encoder and decoder add up while $\mathbf{Q}$ are only from decoder before self-attention.
CrossMerge keeps information individually from each view of full-view slice spatial shuffle block's output in encoder and skip-connects with corresponding view in decoder to maintain view consistence.

For existing CatCrossSkip~\cite{wang2021uformer} operation, it acts as a concatenation-based cross-attention skip-connection which directly projects all information at once, and is rough to lose feature details. While CrossMerge projects the refined feature in individual view.

\section{Experimental Results}
\label{sec:experiments}
In this section, we describe our experimental settings, including the datasets, evaluation metrics, compared methods and implementation details. We first present 
results of compared methods, results for the ablation study of key components,
and then give discussions of limitations and future work.

\subsection{Datasets and Settings}
\subsubsection{3D NCCT Hematoma Dataset}
The 3-D NCCT hematoma private dataset of hemorrhagic stroke is retrospectively obtained from Chinese Intracranial Hemorrhage Image Database (CICHID). The retrospective study was approved by Institutional Review Board of Peking Union Medical College Hospital (Ethics code:S-K1175). 
It's for the challenging hematoma expansion shape dense prediction task, which has 3D image $CT_1$ and its hematoma lesion segmentation $Seg_1$ only at initial time in order to predict the expanded lesion $Seg_2$ at target end time of record in an one-shot setting, where the time interval between initial and target vary widely among different samples.
It's a revised dataset based on previous work~\cite{xiao2021intracerebral} that some failed cases didn't meet the hematoma expansion standard were excluded. 
To evaluate Shuffle-Mixer and selected SOTA backbones upon this image-level lesion growth prediction, we replace 3D U-Net with different backbones in the framework~\cite{xiao2021intracerebral} based on displacement vector field (DVF). We remove clinical data fusion in DVF to evaluate the performance for a fair comparison.
This problem is formulated as a 3D DVF prediction task.
We randomly split our dataset into 212, 30 and 60 volumes for train, validation, and test, respectively. 

\subsubsection{Synapse Dataset}
The synapse (BTCV) public dataset consists of 30 subjects with abdominal CT scans where 13 organs were annotated, which is used as a small dataset to test for vision transformer. Each CT scan 
consists of 80 to 225 slices with 512$\times$512 pixels and slice thickness ranging from 1 to 6 mm. Each volume has been pre-processed by normalizing the intensities in the range of [-1000, 1000] HU to [0, 1]. All images are resampled into the isotropic voxel spacing of 1.0 mm during pre-processing. 
This problem is formulated as a 13 class multi-organ segmentation task.
We split synapse dataset into 24 and 6 for train and test respectively, the same as split setting in~\cite{hatamizadeh2021unetr} for a fair comparison.

\subsubsection{BraTS2019 Dataset}
The BraTS2019 public dataset is provided by the Brain Tumor Segmentation 2019 challenge. Each sample is composed of 4 modalities of brain MRI scans. Each modality has a volume of 240$\times$240$\times$155 which has been aligned into the same space. 
The labels contain 4 classes: background (label 0), necrotic and non-enhancing tumor (label 1), peritumoral edema (label 2) and GD-enhancing tumor (label 4). 
The segmentation accuracy is measured on 3 classes of compositional regions which are enhancing tumor region (ET, label 1), the tumor core region (TC, labels 1 and 4), and the whole tumor region (WT, labels 1,2 and 4). 
It contains 335 cases of patients for offline training and 125 cases for online validation by official server. 

\subsubsection{Evaluation Metrics}
We use dice similarity coefficient (Dice) score as evaluation metric to measure the predicted region accuracy both upon three datasets, represented as: 
$\text { Dice }=\frac{2 * T P}{F P+2 * T P+F N}$.
Additionally, Dice of Change (Doc), Precision, Recall and Jaccard of hematoma expansion region are used to measure predicted region internal filling accuracy more comprehensively upon hematoma dataset, which are formulated as:
$\text{Doc}=\frac{\left|P_{\mathrm{m}}-M_{2}\right| \cap\left|M_{2}-M_{1}\right|}{\left|P_{\mathrm{m}}-M_{2}\right|\cup\left|M_{2}-M_{1}\right|}$,
$\text { Precision }=\frac{T P}{F P+T P}$,
$\text { Recall }=\frac{T P}{T P+F N}$,
$\text { Jaccard }=\frac{T P}{F P+T P+F N}$.
Since shape prediction also concentrates on shape boundary accuracy besides overall region,  Hausdorff Distance 95\% (HD95) is also used to measure the region boundary accuracy.

\subsection{Compared Methods and Implementation Details}
To fully compare different kinds of backbones for dense prediction in medical volume.
For hematoma dataset, 
we use initial hematoma as benchmark, 3D U-Net~\cite{cciccek20163d}, AHNet~\cite{liu20183d}, HyperDensenet~\cite{dolz2018hyperdense} and Att-Unet~\cite{oktay2018attention} as classic simple backbones here. Since it's a one-shot hard task, we also choose cascade ensemble (Cascaded U-Net~\cite{zhao2019recursive}), multi-task ensemble (segResnetVAE~\cite{myronenko20183d}), and model ensemble (nnU-net~\cite{isensee2021nnu}) to compare. Besides, our motivation is to design a strong 3D vision learner leveraging local vision transformer. Thus we compare with benchmark ViT~\cite{dosovitskiy2020image} and SOTA methods of local vision transformer in general vision tasks: Swin-TR~\cite{liu2021swin}, Shuffle-TR~\cite{huang2021shuffle} and CSWin-TR~\cite{dong2021cswin} as transformer-based (TR-based) baselines all for three datasets.
For synapse dataset, besides simple backbones~\cite{milletari2016v} and TR-based baselines, we compare with CNN-TR: TransUNet~\cite{chen2021transunet} and Swin-Unet~\cite{cao2021swin} proposed 
recently and SOTA UNETR~\cite{hatamizadeh2021unetr}.
For BraTS2019 dataset, besides above baselines and previous solutions~\cite{wang20193d,zhao2019bag,valanarasu2020kiu}, we also compare with CNN-TR: TransBTS~\cite{wang2021transbts} proposed 
recently and SOTA Segtran~\cite{li2021medical}.

All networks on hematoma dataset are implemented in Pytorch and MONAI\footnote{https://github.com/Project-MONAI/MONAI} framework with Adam optimization for 200 epochs with batch size 4. The learning rate was set to 0.0005.
All networks on synpase dataset are implemented and evaluated based on the official code\footnote{https://github.com/Project-MONAI/research-contributions/tree/master/UNETR/BTCV} of recent SOTA UNETR. 
All networks on BraTS2019 dataset are implemented and evaluated based on the official code\footnote{https://github.com/askerlee/segtran} of recent SOTA Segtran. The input volume size on three datasets are all set to 128$\times$128$\times$128. In particular, we build our Shuffle-Mixer and other SOTA TR-based methods with the same and common configuration for a fair comparison (see section\ref{appendix:TR configuration} in Supplementary).

\subsection{Comparison with SOTA upon Hematoma Expansion Shape Prediction Task}
\begin{table}[!htbp]
\centering
\caption{Performance comparison upon hematoma expansion shape prediction task. (\textbf{Bold} font is for the best, $\underline{underlined}$ font for the second-best and $\uwave{wavy lines}$ font for the worst. These are apply to all tables in this paper.)}
\label{tb:hematoma-baseline}
\resizebox{0.5\textwidth}{!}{
\begin{tabular}{l|l|cccccc}
\toprule
Type                                 & Model                              & Dice$\uparrow$    & Doc$\uparrow$    & Precision$\uparrow$ & Recall$\uparrow$ & Jaccard$\uparrow$ & HD95$\downarrow$  \\ \midrule
Benchmark                        & origin $Seg_1$                         & 0.5988 & 0      & -         & -      & -       & -     \\ \midrule
\multirow{4}{*}{Simple backbone} & 3D U-Net                           & 0.695  & 0.4486 & 0.6097    & 0.4252 & 0.2994  & 12.12 \\
                                 & AHNet                              & 0.6975 & 0.4565 & 0.6174    & 0.4347 & 0.3065  & 12.02 \\
                                 & HyperDensenet                      & 0.6906 & 0.4299 & \underline{0.6321}    & 0.3837 & 0.2837  & 12.8  \\
                                 & Att-Unet                           & 0.6964 & 0.4495 & 0.6147    & 0.4287 & 0.3     & 12.44 \\ \midrule
\multirow{4}{*}{Ensemble}        & Cascaded U-Net (share) & 0.6938 & 0.4451 & 0.6002    & 0.4263 & 0.2946  & 12.71 \\
                                 & Cascaded U-Net (not share)   & 0.6986 & 0.4574 & 0.606     & 0.4467 & 0.3082  & 11.86 \\
                                 & segResnetVAE                       & 0.6993 & 0.4592 & 0.617     & 0.4383 & 0.3096  & 11.8  \\
                                 & nnU-net                            & \underline{0.7012} & \underline{0.4617} & 0.6171    & 0.4499 & \underline{0.3113}  & \underline{11.78} \\ \midrule
\multirow{6}{*}{TR-based}        & pure ViT                           & \multicolumn{6}{c}{- (Not Converge)}                    \\
                                 & 2D Swin-TR                          & 0.666  & 0.3395 & \textbf{0.6469}    & 0.2629 & 0.2105  & 13.31 \\
                                 & 2D Shuffle-TR                       & 0.6942 & 0.448  & 0.6117    & 0.4228 & 0.2987  & 12.16 \\
                                 & 2D CSWin-TR                         & 0.6898 & 0.434  & 0.6164    & 0.399  & 0.2866  & 12.46 \\
                                 & 3D video Swin-TR                    & 0.6922 & 0.4399 & 0.6165    & 0.4075 & 0.2912  & 12.36 \\
                                 & 3D CSWin-TR                         & 0.6969 & 0.4609 & 0.5939    & \underline{0.4513} & 0.3094  & 11.9  \\ \midrule
Ours                             & 3D Shuffle-Mixer                      & \textbf{0.7067} & \textbf{0.4992} & 0.6013    & \textbf{0.5228} & \textbf{0.3456}  & \textbf{11.05} \\ \bottomrule
\end{tabular}
}
\end{table}

\begin{figure*}
    \centering
    \includegraphics[width=1.0\textwidth, trim={200 210 200 260}, clip]{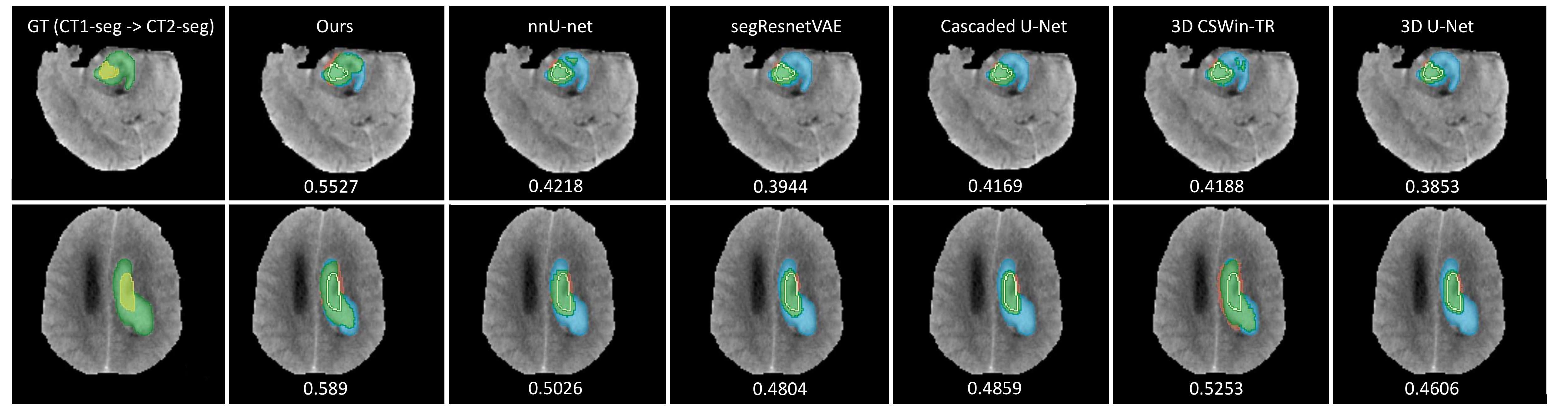}
    \caption{Visualization comparison upon hematoma expansion shape prediction task. The yellow mask in GT denotes $Seg_1$ and green mask denotes $Seg_2$. In prediction results, white line denotes the boundary of $Seg_1$, green mask denotes TP prediction, blue mask denotes FN prediction and orange mask denotes FP prediction. The Doc value of each sample is shown at the bottom of each snapshot.}
    \label{fig:hematoma-vis}
\end{figure*}    
We present the performance upon hematoma expansion shape prediction task in Table~\ref{tb:hematoma-baseline}. Since the initial hematoma region $Seg_1$ of $CT_1$ is directly given in training, it is more important to pay attention to the prediction accuracy of the new expansion region from $Seg_1$ to $Seg_2$ when ensuring the accuracy of the overall region. Therefore, we focus on the major metrics Doc and Jaccard. Our model achieves SOTA performance for all metrics except slightly lower precision. Moreover, it outperforms the second-best model nnU-net by 3.75\% and 3.43\% on two key scores. In particular, our model significantly improved recall by 7.15\% at the expense of 
a little precision sacrificing in contrast to 3D CSWin-TR's second-best recall. In addition, Cascaded U-Net (share) assumes hematoma's linear expansion and Cascaded U-Net (not share) assumes nonlinear expansion. Both mediocre performance indicates that the hematoma doesn't expand regularly with time and our model can make better prediction. All TR-based baselines get poor performance, which indicates that directly applying existing 2D local vision transformer to medical volume doesn't work well, while the Transformer-MLP paradigm designed by us can achieve remarkable results.

We visualize the hematoma expansion shape prediction results in Fig.~\ref{fig:hematoma-vis} of Shuffle-Mixer compared with other selected SOTA methods. Ground truth (GT) shows the expansion are highly irregular, where the hematoma regions of $Seg_2$ and $Seg_1$ are much different. Except for our model, nnU-net and 3D CSWin-TR, the prediction results of others tend to be conservative, with only a slight expansion along the boundary of $Seg_1$. Surprisingly, nnU-net correctly predicts only slightly more expansion region, while 3D CSWin-TR and Shuffle-Mixer correctly predict most of the expansion region, where the angle direction and distance of the predicted expansion region are roughly consistent with GT benefiting from the global receptive field of vision transformer. In contrast to 3D CSWin-TR, FP region of our model is much more smaller.

\subsection{Comparison with SOTA upon Popular 3D Image Segmentation Tasks}
We evaluate our proposed method upon two popular 3D image segmentation tasks in this subsection. The first one is synapse multi-organ segmentation with CT images and the second one is BraTS2019 tumor segmentation with multi-modal MRI images. 

\begin{table*}[!htbp]
\centering
\caption{Performance comparison upon synapse multi-organ CT segmentation task.}
\label{tb:synapse-baseline}
\resizebox{1.0\textwidth}{!}{
\begin{tabular}{l|l|cccccccccccc|c}
\toprule
Type                                 & Model            & Spl    & RKid   & LKid   & Gall   & Eso    & Liv    & Sto    & Aor    & IVC    & Veins  & Pan    & AG     & Average Dice$\uparrow$ \\ \midrule
\multirow{3}{*}{Simple backbone} & V-Net            & \underline{0.9454} & \textbf{0.9354} & \underline{0.9269} & \uwave{0.1521} & \uwave{0.2932} & \underline{0.9633} & \textbf{0.8805} & 0.9004 & \underline{0.8529} & \uwave{0.2881} & \underline{0.7377} & 0.2196 & 0.6746       \\
                                 & 3D U-Net         & 0.8955 & 0.8955 & 0.8937 & \underline{0.5389} & 0.6865 & 0.9424 & 0.7711 & 0.8576 & 0.8108 & 0.6412 & 0.5945 & 0.5674 & 0.7579       \\
                                 & Att-Unet         & 0.8875 & 0.8778 & 0.8758 & \textbf{0.5634} & 0.6948 & \uwave{0.9005} & 0.7805 & 0.837  & 0.8086 & 0.6443 & 0.7064 & 0.6397 & 0.768        \\ \midrule
\multirow{2}{*}{Paper used}      & TransUNet        & 0.9135 & 0.8374 & 0.8332 & 0.445  & \underline{0.7204} & 0.9565 & 0.734  & \textbf{0.9117} & 0.8073 & 0.67   & 0.5927 & 0.5731 & 0.7496       \\
                                 & Swin-Unet        & 0.8751 & 0.908  & 0.9189 & 0.5203 & 0.6028 & 0.9459 & 0.7195 & 0.898  & 0.7986 & \textbf{0.6858} & 0.7218 & 0.5939 & 0.7657       \\ \midrule
Recent SOTA                      & UNETR            & 0.9153 & 0.9184 & 0.9071 & 0.507  & 0.7094 & 0.9482 & 0.8005 & 0.8653 & 0.8117 & 0.6614 & 0.7138 & \underline{0.648}  & \underline{0.7838}       \\ \midrule
\multirow{6}{*}{TR-based}        & pure ViT         & 0.8406 & 0.843  & 0.7915 & 0.2799 & 0.3774 & 0.9155 & \uwave{0.5476} & 0.7646 & \uwave{0.603}  & 0.3967 & \uwave{0.312}  & \uwave{0.1565} & \uwave{0.569}        \\
                                 & 2D Swin-TR        & \uwave{0.833}  & \uwave{0.8234} & \uwave{0.783}  & 0.349  & 0.5269 & 0.936  & 0.6389 & \uwave{0.7072} & 0.672  & 0.3746 & 0.4817 & 0.2686 & 0.6162       \\
                                 & 2D shuffle-TR     & 0.8598 & 0.8842 & 0.8827 & 0.4405 & 0.667  & 0.9516 & 0.7263 & 0.9003 & 0.8357 & 0.6795 & 0.6221 & 0.5417 & 0.7493       \\
                                 & 2D CSWin-TR       & 0.8431 & 0.8797 & 0.8725 & 0.4483 & 0.6756 & 0.9439 & 0.6981 & 0.8993 & 0.8363 & \underline{0.6803} & 0.6185 & 0.5409 & 0.7447       \\
                                 & 3D video Swin-TR  & 0.9102 & 0.9112 & 0.8682 & 0.5381 & 0.5993 & 0.9534 & 0.7117 & 0.8317 & 0.7493 & 0.6068 & 0.576  & 0.5139 & 0.7308       \\
                                 & 3D CSWin-TR       & 0.9168 & 0.9127 & 0.8985 & 0.4143 & 0.6589 & 0.9461 & 0.7684 & 0.8755 & 0.8184 & 0.6083 & 0.6845 & 0.5841 & 0.7572       \\ \midrule
Ours                             & 3D Shuffle-Mixer    & \textbf{0.9478} & \underline{0.9353} & \textbf{0.9282} & 0.5153 & \textbf{0.7253} & \textbf{0.9667} & \underline{0.8314} & \underline{0.9015} & \textbf{0.8597} & 0.6772 & \textbf{0.7719} & \textbf{0.6763} & \textbf{0.8114}  \\ \midrule
\end{tabular}
}
\end{table*}

\begin{figure*}
    \centering
    \includegraphics[width=1.0\textwidth, trim={330 100 330 150}, clip]{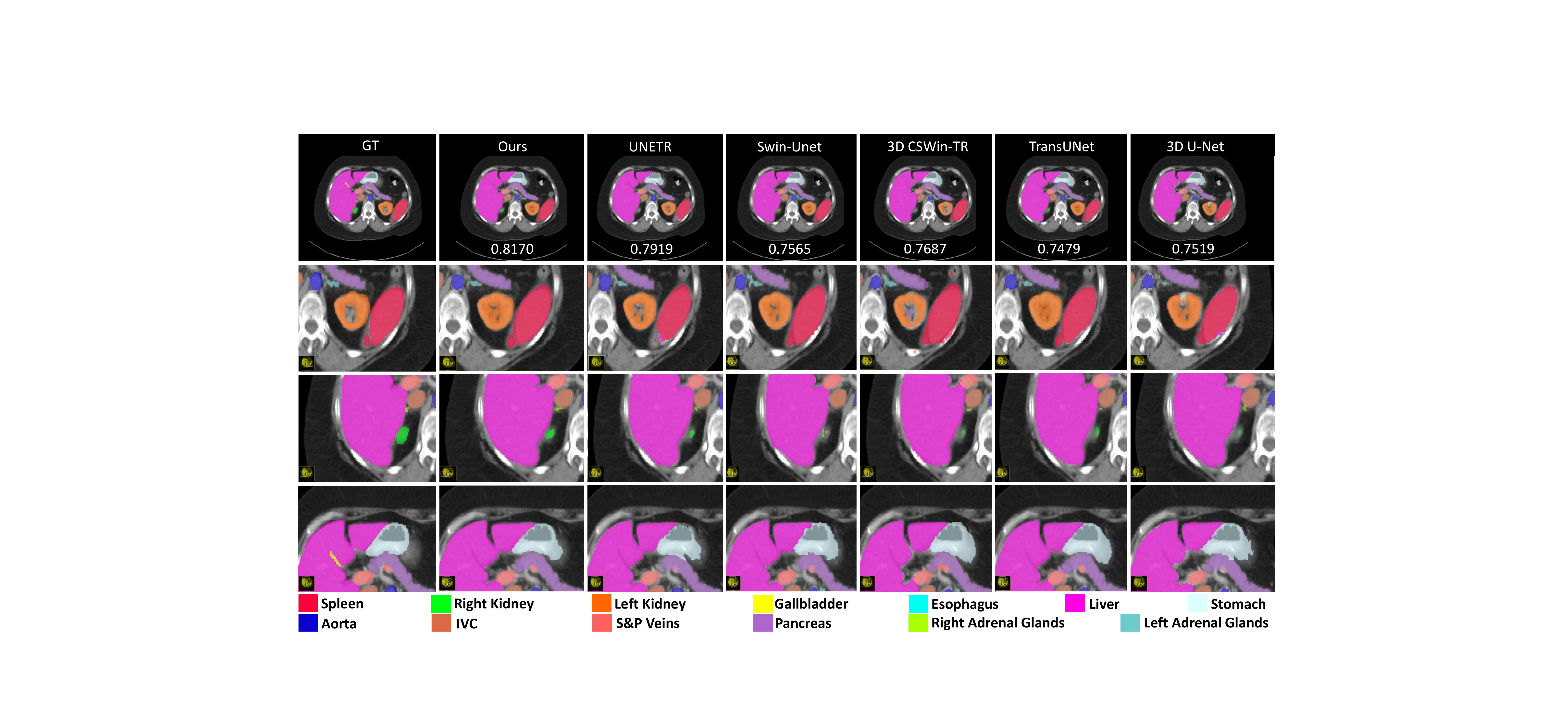}
    \caption{Visualization comparison upon synapse multi-organ CT segmentation task. Snapshots in the first row are overall visualizations and the last three lines are zoom-in region visualizations. The average dice value of each sample is shown at the bottom of each snapshot in the first row.}
    \label{fig:synapse-vis}
\end{figure*} 

\textbf{Synapse Multi-organ Segmentation in CT Images:} We present the performance upon synapse multi-organ CT segmentation task in Table~\ref{tb:synapse-baseline}. Among all the 12 classes of organ, Shuffle-Mixer achieves the best performance in 7 classes, second-best performance in 3 classes, and good results in the remaining two classes, with 2.76\% better than the second-best model UNETR on average dice score.

We visualize the synapse multi-organ CT segmentation results in Fig.~\ref{fig:synapse-vis} between Shuffle-Mixer and other selected SOTA methods for visualization comparison. From the overall visualization and three zoom-in regions, it can be seen that the prediction results of our model are more precise, refined and closer to GT.

\begin{table}[!htbp]
\centering
\caption{Performance comparison upon BraTS2019 multi-modal brain tumor MRI segmentation task.}
\label{tb:brats-baseline}
\resizebox{0.5\textwidth}{!}{
\begin{tabular}{l|l|ccc|c}
\toprule
Type                                 & Model                & ET     & WT     & TC     & Average Dice$\uparrow$ \\
                                 \midrule
\multirow{3}{*}{Simple backbone} & V-Net                & 0.6362 & 0.8583 & 0.653  & 0.7158       \\
                                 & 3D U-Net             & 0.6165 & 0.8404 & 0.6675 & 0.7082       \\
                                 & Att-Unet             & 0.7148 & 0.8885 & 0.8099 & 0.8044       \\
                                 \midrule
\multirow{4}{*}{Paper used}      & Extension of nnU-Net & 0.737  & 0.894  & 0.807  & 0.8127       \\
                                 & Bag of tricks        & 0.729  & \uline{0.904}  & 0.802  & 0.8117       \\
                                 & KiU-Net              & 0.734  & 0.877  & 0.742  & 0.7843       \\
                                 & TransBTS             & 0.735  & 0.895  & 0.803  & 0.811        \\
                                 \midrule
Recent SOTA                      & Segtran              & \uline{0.7381} & 0.8914 & \uline{0.8122} & \uline{0.8139}       \\
\midrule
\multirow{6}{*}{TR-based}        & pure ViT             & 0.583  & 0.822  & 0.608  & 0.671        \\
                                 & 2D Swin-TR            & 0.6392 & 0.8336 & 0.7283 & 0.7337       \\
                                 & 2D Shuffle-TR         & 0.6857 & 0.8775 & 0.7477 & 0.7703       \\
                                 & 2D CSWin-TR           & 0.6723 & 0.8648 & 0.7391 & 0.7587       \\
                                 & 3D video Swin-TR      & 0.6862 & 0.8621 & 0.744  & 0.7641       \\
                                 & 3D CSWin-TR           & 0.6869 & 0.8611 & 0.7531 & 0.767        \\
                                 \midrule
Ours                             & 3D Shuffle-Mixer        & \textbf{0.7482} & \textbf{0.9064} & \textbf{0.8147} & \textbf{0.8231}      \\
\bottomrule
\end{tabular}
}
\end{table}

\textbf{BraTS2019 Tumor Segmentation in MRI Images:} We present the performance upon BraTS2019 multi-modal brain tumor MRI segmentation task in Table~\ref{tb:brats-baseline}. 
Shuffle-Mixer achieves SOTA results on three classes of tumor regions with 0.92\% better than the second-best model Segtran on average dice score. Since the validation samples are not public, so we omit visualization on BraTS2019 here.

\subsection{Effectiveness of Key Component}
\begin{table*}[!htbp]
\centering
\caption{Overall ablation study. Effectiveness of each key component of our proposed method upon three datasets. (\color{blue}{Blue} \color{black}{superscript indicates the performance with best baselines and} \color{red}{red} \color{black}{superscript indicates the performance with ablation (1) basic pure 3D Shuffle-Mixer.})}
\label{tb:model-ablation}
\resizebox{1.0\textwidth}{!}{
\begin{tabular}{lll|cc|c|ccc|c}
\toprule
\multicolumn{3}{l|}{\multirow{2}{*}{Overall method form}}             & \multicolumn{2}{c|}{Hematoma} & Synapse      & \multicolumn{4}{c}{BraTS2019}           \\ \cline{4-10}
\multicolumn{3}{l|}{}                                        & Doc$\uparrow$          & Jaccard$\uparrow$       & Average Dice$\uparrow$ & ET     & WT     & TC     & Average Dice$\uparrow$ \\ \midrule
\multicolumn{3}{l|}{Best baselines (nnU-net, UNETR, Segtran)} & \color{blue}{0.4617}       & \color{blue}{0.3113}        & \color{blue}{0.7838}       & {0.7381} & {0.8914} & {0.8122} & \color{blue}{0.8139}              \\ \midrule
\multicolumn{1}{l|}{3D Shuffle-Mixer}  & \multicolumn{1}{l|}{ASES}  & CrossMerge  & \multicolumn{7}{c}{}             \\ \midrule
\multicolumn{1}{c|}{\checkmark}                 &      \multicolumn{1}{c|}{}                     &              & \color{red}{0.4878}$^{\color{blue}{+2.61\%}}$       & \color{red}{0.3347}$^{\color{blue}{+2.34\%}}$        & \color{red}{0.8018}$^{\color{blue}{+1.80\%}}$       & {0.7443} & {0.9031} & {0.8109} & \color{red}{0.8194}$^{\color{blue}{+0.55\%}}$       \\
\multicolumn{1}{c|}{\checkmark}                 & \multicolumn{1}{c|}{\checkmark}                         &            & 0.4965$^{\color{red}{+0.87\%}}$       & 0.3431$^{\color{red}{+0.84\%}}$        & 0.8096$^{\color{red}{+0.78\%}}$       & 0.7475 & 0.906  & 0.8134 & 0.8223$^{\color{red}{+0.29\%}}$       \\
\multicolumn{1}{c|}{\checkmark}                 &    \multicolumn{1}{c|}{}                       & \multicolumn{1}{c|}{\checkmark}           & 0.4912$^{\color{red}{+0.34\%}}$       & 0.3379$^{\color{red}{+0.32\%}}$        & 0.8033$^{\color{red}{+0.15\%}}$       & 0.7453 & 0.9037 & 0.812  & 0.8203$^{\color{red}{+0.09\%}}$        \\
\multicolumn{1}{c|}{\checkmark}                 & \multicolumn{1}{c|}{\checkmark}                         & \multicolumn{1}{c|}{\checkmark}          & \textbf{0.4992}$^{\color{blue}{+3.75\%},\color{red}{1.14\%}}$       & \textbf{0.3456}$^{\color{blue}{+3.43\%},\color{red}{1.09\%}}$        & \textbf{0.8114}$^{\color{blue}{+2.76\%},\color{red}{0.96\%}}$       & 0.7482 & 0.9064 & 0.8147 & \textbf{0.8231}$^{\color{blue}{+0.92\%},\color{red}{0.37\%}}$      \\
\bottomrule
\end{tabular}
}
\end{table*}
3D Shuffle-Mixer's full version consists of three key components: 3D Shuffle-Mixer, ASES and CrossMerge. To evaluate the effectiveness and importance of each key component, we conduct the ablation study which compares full version of our model with other three degraded versions as ablation variants: (1) The simplest version which only use pure 3D Shuffle-Mixer as first basic variant; (2) We use simple 3D Shuffle-Mixer equipped with ASES as second variant; (3) We use simple 3D Shuffle-Mixer with CrossMerge as third variant; (4) The full version consists of all three key components. 

Table~\ref{tb:model-ablation} shows the overall ablation results. The simplest version (1) is already better than the best baseline on three datasets significantly, which outperforms by 2.61\% and 2.34\% on Doc and Jaccard score upon hematoma task, by 1.80\% and 0.55\% on average dice upon other two segmentation tasks as well. It shows the basic structure of our Shuffle-Mixer has universal superiority among three dense prediction tasks. Ablation (2) equipping with ASES futher improves the performance by approximately 0.8\% to 0.9\% upon hematoma and synapse tasks and 0.3\% on BraTS2019. This is because ASES rectifies and enhances the decoupled features on spatial and channel dimensions adaptively, which is also a significant improvement. Ablation (3) adpoting CrossMerge improves the basis by 0.34\% and 0.32\% upon hematoma task, by approximately 0.1\% to 0.2\% upon two segmentation tasks. Although CrossMerge improves not much as the first two key components, overall stable improvement also verifies the effectiveness of the skip-connection method proposed by us, which is more suitable for pyramid local vision transformer to add the missing multi-resolution details with full-view structure. The full version (4) of our method achieves the best performance along all ablations and surpasses best baselines on three datasets by a large margin. All ablation study mentioned above demonstrates that each key component plays a role in the overall architecture and improves the performance.

\subsection{Discussion of Limitations and Future Work}
3D Shuffle-Mixer is proposed specially for dense prediction in medical volume. In order to better deal with 3D volumetric data, we reduce computational complexity as much as possible while ensuring performance via concise but efficient computation (e.g. W-MSA and sparse axial-MLP), parameter sharing and full-view 2D computing. Our model not only possess the learning capability of vision transformer on large amounts of data revealed in the general vision tasks, so as to improve the upper limit of performance, but also has greater inductive bias power of the local vision transformer for dense prediction due to the full-view spatial shuffle and context mixing. 
The data efficient learning ability from small dataset is demonstrated in our experiments upon synapse dataset. Although our method achieves excellent performance, it comes with limitations. One limitation is that there're some other dense prediction tasks, e.g. 
reconstruction, denoising, synthesis, etc, which haven't been fully verified on. In the future, we will investigate the advantages of our model of Local Vision Transformer-MLP paradigm upon other various downstream dense prediction tasks to explore better performance and improvements of Shuffle-Mixer. Another potential research direction is to extend our method to self-supervised learning. Since advanced vision transformer has proven excellent for self-supervision~\cite{caron2021emerging,bao2021beit} in general vision tasks due to its more unique pretext task forms which contribute to the strong feature representation ability.

Another limitation is that all dimensions of 3D volume are the same size in our setting for convenience, where processing of views in different sizes requires additional pre-processing or hyperparameter presetting in Shuffle-Mixer network to facilitate feature fusion learning from full-view. In the future, we will explore more appropriate ways to handle different sizes of full-view information. Besides, due to task setting, we have no chance to investigate knowledge transferability with transfer learning in this paper, we'll explore generalization ability via cross-domain evaluation.

\section{Conclusion}
\label{sec:conclusion}
In this paper, we have proposed a novel 3D Shuffle-Mixer network exploiting Local Vision Transformer-MLP paradigm for dense prediction in medical volume. 
In particularly, our model rearranges volume into full-view slices to obtain redundant spatial information, mixes slice-aware features and aggregates view-aware features to fully capture the volume context in a context-aware manner. Besides, we have proposed an Adaptive Scaled Enhanced Shortcut to rectificate and augment feature representation of window-based local vision transformer. Furthermore, an appropriate skip-connection form CrossMerge for pyramid transformer is presented to merge feature from encoder to decoder. The experimental results and analysis upon dense prediction tasks have demonstrated that our proposed model outperforms other SOTA compared methods in both quantitative and qualitative measures.

\normalem
\bibliographystyle{IEEEtran}
\bibliography{references} 

\clearpage
\appendices
\section*{Supplementary}
\subsection{Inner-Analysis upon 3D Shuffle-Mixer}
\label{Appendix:ia-sm}
\begin{table}[!htbp]
\centering
\caption{Inner ablation study. Effectiveness of each composition of 3D Shuffle-Mixer upon three datasets.}
\label{tb:sm-ablation}
\resizebox{0.5\textwidth}{!}{
\begin{tabular}{l|c|c|c}
\toprule
\multirow{2}{*}{3D Shuffle-Mixer form} & Hematoma & Synapse      & BraTS2019    \\  \cline{2-4} 
                                       & Doc$\uparrow$      & Average Dice$\uparrow$ & Average Dice$\uparrow$ \\ \midrule
Ours (full version)                    & \textbf{0.4992}   & \textbf{0.8114}       & \textbf{0.8231}       \\ \midrule
W-MSA w/o shuffle                      & 0.4723   & 0.7868       & 0.7841       \\
Single view (only H,W)                 & 0.4921   & 0.7994       & 0.8179       \\ \midrule
W/o context mixing                     & 0.4916   & 0.7945       & 0.8152       \\
MLP dense context mixing               & 0.4887   & 0.7982       & 0.8167       \\
MSA dense context mixing               & \multicolumn{3}{c}{- (Exceed GPU)}         \\ \midrule
Mixer-Shuffle                          & 0.489    & 0.8085       & 0.8154       \\ \midrule
Not slice-aware                        & 0.4881   & 0.8007       & 0.8155       \\
Not view-aware                         & 0.4865   & 0.8038       & 0.8173      \\
\bottomrule
\end{tabular}
}
\end{table}
To have a better understanding and inner-analysis of basic 3D Shuffle-Mixer, we conduct a series of extended experiments on the composition form upon 3D Shuffle-Mixer. To facilitate comparison, we use the full version of our method as initial structure for all ablations and subsequent settings were limited to the adjustments on the basic 3D Shuffle-Mixer part in this section, and the other two key components remained same using the optimized structure for a fair comparison. The same way compared in the section\ref{sec:ia-ases} and \ref{sec:ia-cm}.

Table~\ref{tb:sm-ablation} shows inner ablation study results of 3D Shuffle-Mixer: (1) To investigate the effectiveness of the full-view slice spatial shuffle operation, we compare with two ablations, where the first one removes spatial transpose shuffle turning into a pure W-MSA and the second one only reserves the $(H,W)$ single view. The two ablations degrade the performance seriously on all datasets since spatial shuffle helps W-MSA learn from cross-windows to increase receptive field for transformer, and single-view misses redundant volume information from other views. (2) To investigate the effectiveness of the slice-aware volume context mixing,  we compare with three ablations, where the first one removes the mixing, the latter two replace axial-MLP along the third dimension $st$ and channel dimension $sc$ by MLP dense mixing and MSA dense mixing for the whole $(st, sc)$. W/o mixing and MLP dense mixing both degrade performance since mixing brings into missing volume context and axial-MLP is sparse which mitigates over-fitting, however, dense MLP introduces extra parameters and increases the risk of MLP overfitting, which leads to poor context processing. MSA dense mixing directly exceeds GPU memory and becomes uncomputable. (3) To investigate the order of Shuffle and Mixer, we compare with one ablation where Mixer is at first and shuffle at last. Performance reduction proves the logic of Shuffle-Mixer learning full-view slice spatial context first and then remaining slice-aware volume context is correct and in line with expectations. (4) To investigate the importance of slice-aware and view-aware, we compare with two ablations to remove APE turning into slice or view not aware. Dramatic reduction indicates that the capacity of two awareness is necessary for the module to correctly distinguish which slice or view the information comes from in the process of mixing or aggregate, and realize the importance of different context.

\subsection{Inner-Analysis upon Adaptive Scaled Shortcut}
\label{sec:ia-ases}
\begin{table}[!htbp]
\centering
\caption{Inner ablation study. Effectiveness of each composition of adaptive scaled enhanced shortcut upon three datasets.}
\label{tb:ASES-ablation}
\resizebox{0.5\textwidth}{!}{
\begin{tabular}{l|c|c|c}
\toprule
\multirow{2}{*}{ASES form} & Hematoma & Synapse      & BraTS2019    \\ \cline{2-4} 
                                               & Doc$\uparrow$      & Average Dice$\uparrow$ & Average Dice$\uparrow$ \\ \midrule
W/o ASES                                        & 0.4912   & 0.8033       & 0.8203       \\
Only spatial enhance on W-MSA                  & 0.4978   & 0.8082       & 0.8219       \\
Only channel enhance on MLP                    & 0.4937   & 0.8065       & 0.8211       \\
Ours (full version)                            & \textbf{0.4992}   & \textbf{0.8114}       & \textbf{0.8231}       \\ \midrule
3D CSWin-TR                                    & 0.4609   & 0.7572       & 0.767        \\
3D CSWin-TR w/ ASES                             & 0.4674   & 0.7681       & 0.773        \\ \midrule
Channel enhance-spatial enhance                & 0.4913   & 0.8008       & 0.8175      \\
\bottomrule
\end{tabular}
}
\end{table}
To have a better understanding and inner-analysis of ASES, we conduct a series of extended experiments on the composition form upon ASES. 

Table~\ref{tb:ASES-ablation} shows inner ablation study results of ASES: (1) To investigate the effectiveness of the spatial enhance and channel enhance in ASES, we compare with three ablations, where the first one removes ASES, the second one only has spatial enhance on W-MSA and the third one only has channel enhance on MLP. W/o ASES achieves the lowest performance, adding spatial or channel enhance alone both improves performance to some extent on all three datasets, and improves the most with both enhanced shortcuts since decoupling spatial and channel features are rectified and augmented successively. (2) To investigate the general enhanced capability of ASES, we compare with two ablations, where one is pure 3D CSWin-TR and another one is 3D CSWin-TR w/ ASES. We find equipping with ASES strongly improves performance on other local vision transformer as well which demonstrates the general capability and suitability of our ASES. (3) To investigate the order of spatial and channel enhanced shortcuts,  we  compare  with  one  ablation  where channel enhance is at first and spatial enhance at last. The performance degradation shows that spatial enhance needs to be applied to W-MSA and channel enhance needs to be applied to MLP, which are consistent with vision transformer design and our intuition.

\subsection{Inner-Analysis upon CrossMerge Skip-Connection}
\label{sec:ia-cm}
\begin{table}[!htbp]
\centering
\caption{Inner ablation study. Effectiveness of each composition of skip-connection upon three datasets.}
\label{tb:CrossMerge-ablation}
\resizebox{0.5\textwidth}{!}{
\begin{tabular}{l|c|c|c}
\toprule
\multirow{2}{*}{Skip-Connection form} & Hematoma & Synapse      & BraTS2019    \\ \cline{2-4} 
                                      & Doc$\uparrow$      & Average Dice$\uparrow$ & Average Dice$\uparrow$ \\ \midrule
CatLinear (basic U-Net)                & 0.4878   & 0.8018       & 0.8194       \\
CatSkip (Uformer)                     & 0.4945   & 0.808        & 0.8202       \\
CrossSkip (Uformer)                   & 0.4917   & 0.8023       & 0.8187       \\
CatCrossSkip (Uformer)                & 0.4938   & 0.8076       & 0.821        \\
CrossMerge (Ours)                     & \textbf{0.4992}   & \textbf{0.8114}       & \textbf{0.8231}      \\
\bottomrule
\end{tabular}
}
\end{table}
To find a skip-connection form suitable for vision transformer and have a better understanding of CrossMerge, we conduct extended experiments as five ablations on the composition form upon skip-connections. Table~\ref{tb:CrossMerge-ablation} shows inner ablation study results of skip-connection: CatLinear is used in basic U-Net. CatSkip, CrossSkip and CatCrossSkip are three variants proposed in Uformer~\cite{wang2021uformer}. CrossMerge is proposed by us. Our CrossMerge achieves the best performance on three datasets, performances of CatLinear and CrossSkip are generally poor since CatLinear is more suitable for backbones of CNN structure and CrossSkip directly converts features from encoder into key, value and converts from decoder into query, missing the merge procedure of information. CatSkip and CatCrossSkip achieve the medium performance. Although these two forms merge features from encoder to decoder, our CrossMerge benefits from refined view-wise skip-connections and performs more stable due to PWConv2D.

\subsection{TR-based Method Build Configurations}
\label{appendix:TR configuration}
\begin{table*}[!htbp]
\centering
\caption{Detailed Build configurations for TR-based Method Implementation.}
\label{tb:TR-based configuration}
\resizebox{1.0\textwidth}{!}{
\begin{tabular}{c|c|c|c|c|c|c}
\toprule
Depth   & \#Token Resolution & \#Window Size & \#Embedding Channels & \#Block Numbers & \#Head Numbers & \#MLP Ratio \\ \midrule
Stage-1 & $\frac{H}{4} \times \frac{W}{4} \times \frac{D}{4}$                & 4             & 96                   & 1               & 3              & 4           \\ \midrule
Stage-2 & $\frac{H}{8} \times \frac{W}{8} \times \frac{D}{8}$                & 4             & 192                  & 2               & 6              & 4           \\ \midrule
Stage-3 & $\frac{H}{16} \times \frac{W}{16} \times \frac{D}{16}$                & 4             & 384                  & 8               & 12             & 4           \\ \midrule
Stage-4 & $\frac{H}{32} \times \frac{W}{32} \times \frac{D}{32}$                & 4             & 768                  & 1               & 24             & 4           \\ 
\bottomrule
\end{tabular}
}
\end{table*}
 We refer to the common configuration in general vision task works~\cite{liu2021swin,huang2021shuffle,dong2021cswin,liu2021video} to build Shuffle-Mixer and other TR-based compared methods. Table~\ref{tb:TR-based configuration} shows the detailed build configurations for TR-based methods implemented in this paper. The depth of pyramid architecture is set as 4, that is, stage-4 is the bottom stage in encoder and setting in decoder is symmetric. Token resolution in stage-1 is $\frac{H}{4} \times \frac{W}{4} \times \frac{D}{4}$ through patch embedding and 2$\times$ downsampling of resolution in each stage. The window size is set as 4 in all stages. Block numbers are set as \{1, 2, 8, 1\} and head numbers are set as \{3, 6, 12, 24\} in each stage. Embedding channel dimensions are set as \{96, 192, 384, 768\} which are 32$\times$ \#head numbers. The expansion ratio of each MLP is set as 4 for all experiments. Besides, the linear layer in vision transformer's MLP of Shuffle-Mixer is implemented by PWConv2D for consideration of better handling 2D spatial slices.

\subsection{Analysis upon Computation Complexity}
We use W-MSA in spatial shuffle block for efficient global context modeling with low computation complexity compared with pure MSA. Supposing each window contains $M \times M$ tokens, the computational complexity of pure MSA and W-MSA on a spatial slice of $H \times W$ tokens with channel dimension $C$ is:
\begin{equation}
\begin{aligned}
&\Omega(\mathbf{Pure\text{ }MSA})=4 H W C^{2}+2 (H W)^{2} C \\
&\Omega(\mathbf{W\text{-}MSA})=4 H W C^{2}+2 M^{2} H W C \\
\end{aligned}
\end{equation}
The complexity of pure MSA is quadratic of slice spatial $(H,W)$, yet W-MSA reduces complexity to linear with fixed window size $M$ ($M=4$, which can be regarded as a constant term) by computing self-attention in each non-overlapping local window. Generally, $M^{2} \ll HW$ where $HW$ is at huge resolution in dense prediction.

Besides, we use axial-MLP in mixing with sparse computation to capture remaining volume context. We compare the complexity of axial-MLP mixing ($\mathbf{A\text{-}MLP\text{-}M}$) with dense MLP context mixing ($\mathbf{D\text{-}MLP\text{-}M}$) and dense MSA context mixing ($\mathbf{D\text{-}MSA\text{-}M}$) introduced in section\ref{Appendix:ia-sm} as follows:
\begin{equation}
\begin{aligned}
&\Omega(\mathbf{D\text{-}MSA\text{-}M})=8 H W D C^{2}+2 (H W D)^{2} C + 2 \alpha H W D C^{2} \\
&\Omega(\mathbf{D\text{-}MLP\text{-}M})=2 \alpha H W D C (D C)+ 3 H W D C^{2} \\
&\Omega(\mathbf{A\text{-}MLP\text{-}M})=2 \alpha H W D C (D+C)+ 3 H W D C^{2} \\
\end{aligned}
\end{equation}
Noticed that $\mathbf{D\text{-}MSA\text{-}M}$ is of pure 3D usage here which is quadratic of 3D volume $(H,W,D)$ with a heavy computation. $\mathbf{A\text{-}MLP\text{-}M}$ uses the residual two-branch structure to project context sparsely and individually in order to mitigate over-fitting and acquire better performance, which costs a relative low computation in contrast to $\mathbf{D\text{-}MLP\text{-}M}$ where $D+C \ll DC$. Consider the complexity of view MLP aggregator as $3 H W D C^{2}$, which can be negligible as a small computational term over whole pipeline of Shuffle-Mixer block ($\mathbf{SM\text{-}B}$), therefore the complexity of $\mathbf{SM\text{-}B}$ is computed as:
\begin{equation}
\Omega(\mathbf{SM\text{-}B})=\Omega(\mathbf{W\text{-}MSA})+\Omega(\mathbf{A\text{-}MLP\text{-}M})
\end{equation}
Moreover, the W-MSA and axial-MLP are parameters shared among full-view which further reduces complexity dramatically and enhances locality with inductive bias.

\begin{table}[!htbp]
\centering
\caption{Model Complexity Comparison.}
\label{tb:complexity}
\resizebox{0.5\textwidth}{!}{
\begin{tabular}{l|c|c|c}
\toprule
Type                                      & Selected Model            & \#Param. ($10^{6}$)$\downarrow$           & FLOPs ($10^{10}$)$\downarrow$            \\ \midrule
\multirow{2}{*}{2D CNN-TR}            & Swin-Unet        & 41.35            & 36.10            \\
                                      & TransUNet        & 66.88            & 104.12           \\ \midrule
\multirow{3}{*}{3D CNN-TR}            & TransBTS         & 32.99            & 26.42            \\
                                      & Segtran          & \textbf{32.97}            & 65.37            \\
                                      & UNETR            & 95.83            & \textbf{17.45}            \\ \midrule
\multirow{4}{*}{3D TR block} & pure ViT         & \uwave{149.64}           & \uwave{993.47}           \\
                                      & 3D video Swin-TR & 60.68            & 22.87            \\
                                      & 3D CSWin-TR      & 89.56            & 38.41            \\
                                      & Shuffle-Mixer    & 85.18 (6th-lowest) & 32.30 (4th-lowest) \\ 
                                      \bottomrule
\end{tabular}
}
\end{table}
Tabel~\ref{tb:complexity} shows the results of model complexity. We select some recent SOTA works and classify them into the 2D CNN-TR hybrid model, 3D CNN-TR hybrid model, and 3D TR designed block model for complexity comparison. All models are tested under the setting of hematoma shape prediction task. Although 2D CNN-TRs process 2D convolution on image scans, they use pure 3D MSA directly in the model which leads to a huge computation increase. Therefore, pure 3D ViT has the largest number of parameters and FLOPs, far more than any other models, which makes running pyramid pure ViT out of memory or extremely slow. In addition, Segtran has the optimal number of parameters and UNETR has the optimal FlOPs. Shuffle-Mixer has the 6th-lowest parameters (85.16 million) and the 4th-lowest FLOPs (32.30 billion) among all nine models which is an acceptable and moderate computational complexity compared to UNETR (95.83 million \#Params.) and Swin-Unet (36.10 billion FLOPs). Compared with 3D CSWin-TR, Shuffle-Mixer has less complexity and achieves SOTA performance on 3D medical volume.

\subsection{Robustness and Convergence Speed Analysis}
\begin{table}[!htbp]
\centering
\caption{Standard deviation comparison of performance from 4 runs for each model upon two datasets.}
\label{tb:robustness}
\resizebox{0.5\textwidth}{!}{
\begin{tabular}{l|c|l|c}
\toprule
\multicolumn{2}{c|}{Hematoma}              & \multicolumn{2}{c}{Synapse}      \\ \midrule
Selected Model          & STDEV.(\%) & Selected Model & STDEV.(\%) \\ \midrule
3D U-Net                 & 0.8292       & 3D U-Net        & 1.4917       \\
Cascaded U-Net (not share) & 0.4869       & TransUNet       & 1.1031       \\
segResnetVAE             & 0.7065       & Swin-Unet       & 1.1769       \\
nnU-net                  & 0.4356       & UNETR           & 1.1790       \\
3D CSWin-TR              & 0.3925       & 3D CSWin-TR     & 0.7928       \\
Ours                     & \textbf{0.1717}       & Ours            & \textbf{0.4211}      \\
\bottomrule
\end{tabular}
}
\end{table}
In addition to the good performance of Shuffle-Mixer upon three dense prediction datasets of 3D medical image analysis, we find our model has remarkable robustness and competitive convergence speed. Table~\ref{tb:robustness} shows the standard deviation (STDEV) performance from each 4 runs of select models upon hematoma and synapse datasets. It can be easily seen that the model with pure CNN or hybird CNN-TR structure is more unstable than the 3D CSWin-TR and Shuffle-Mixer with 3D designed TR block. 3D U-Net's STDEV reaches at 0.8292\% and 1.4917\% respectively which is the most unstable. The more vision transformer components are added, the more stable the model will be, which is also consistent with empirical knowledge~\cite{paul2021vision} in general vision tasks. Our Shuffle-Mixer achieves the minimum STDEV at 0.1717\% and 0.4211\% both upon two datasets which has remarkable robustness among selected models.

\begin{figure}
    \centering
    \includegraphics[width=0.5\textwidth, trim={270 150 260 150}, clip]{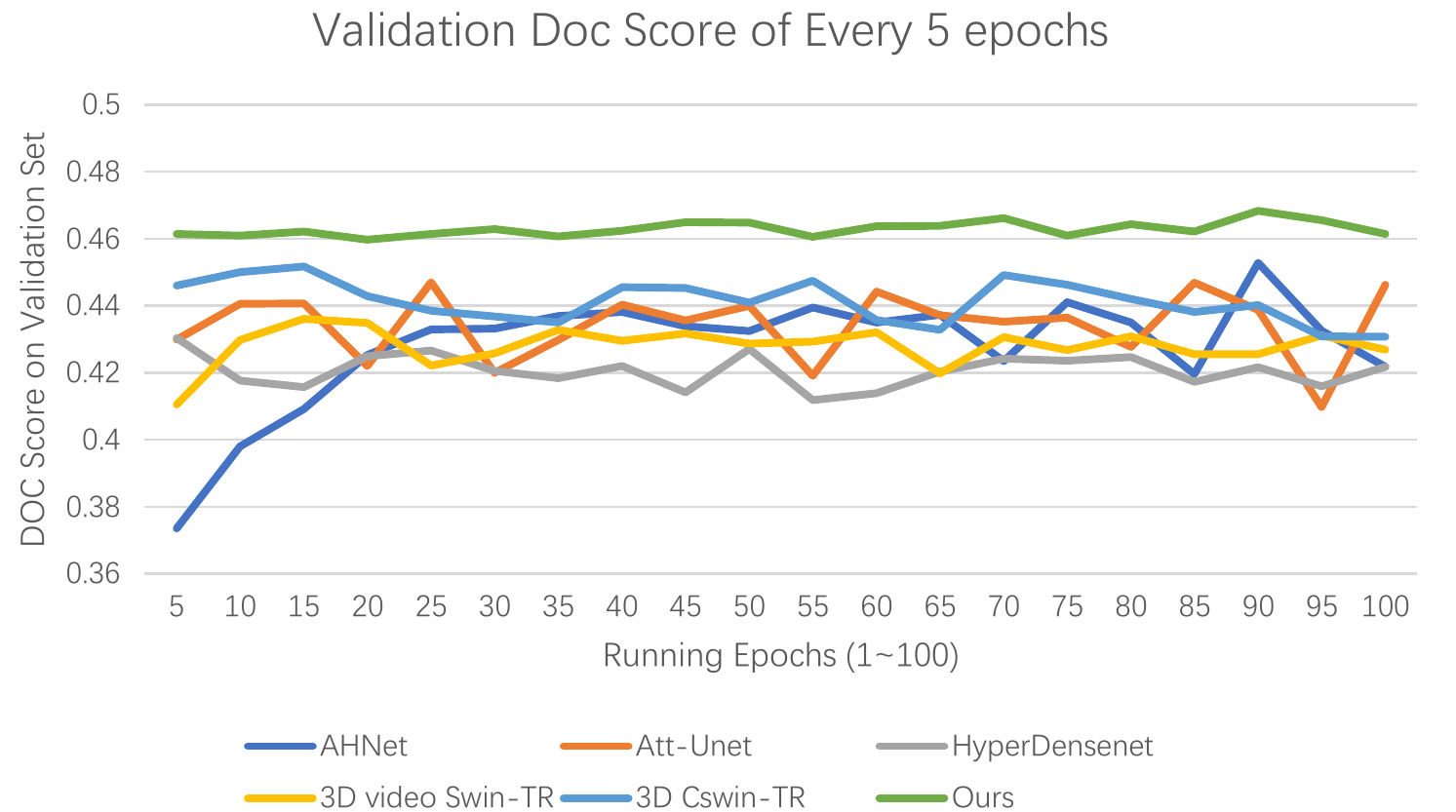}
    \caption{Doc score of every 5 epochs on validation set upon hematoma expansion shape prediction task.}
    \label{fig:hematomaline-vis}
\end{figure}    

\begin{figure}
    \centering
    \includegraphics[width=0.5\textwidth, trim={200 110 200 110}, clip]{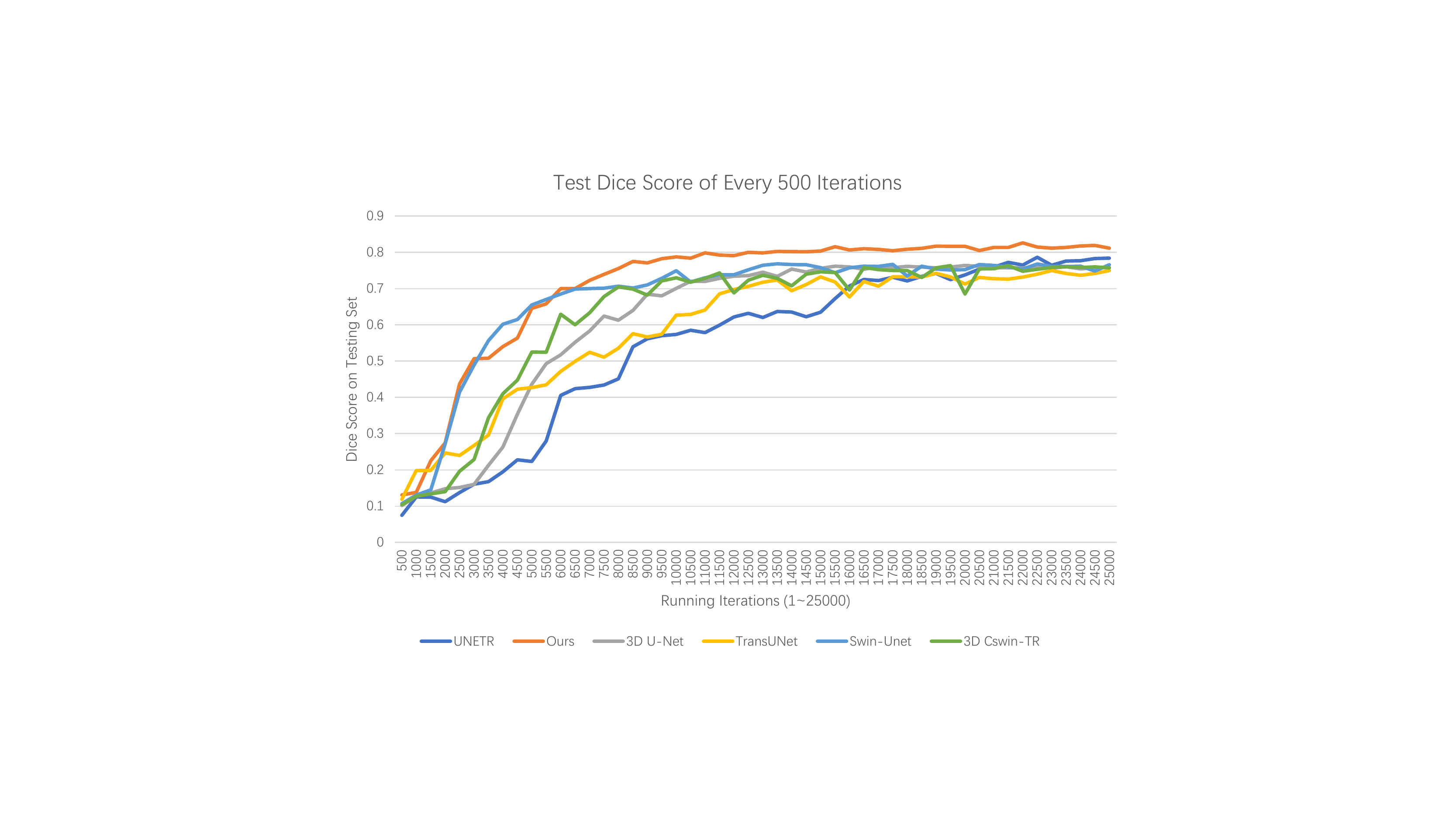}
    \caption{Dice score of every 500 iterations on testing set upon synapse multi-organ CT segmentation task.}
    \label{fig:synapseline-vis}
\end{figure}   
We also compare the model convergence speed during training to analyze robustness. Fig.~\ref{fig:hematomaline-vis} shows the Doc score of every 5 epochs on validation set upon hematoma expansion shape prediction task. Shuffle-Mixer's Doc is significantly higher than other models, and fluctuates only in a small range of 0.46 to 0.47, where other models fluctuate more severely. Shuffle-Mixer has performed better than other models since the first epoch, and stays ahead throughout the whole training process which shows our model 
converges steadily almost within the first few epochs. Fig.~\ref{fig:synapseline-vis} shows the dice score of every 500 iterations on testing set upon synapse multi-organ CT segmentation task since there is no validation set for it. It can be seen that only the performance of Shuffle-Mixer and Swin-Unet grows at the fastest speed as training iterations increase, where they quickly convergence and the performance tends to be stable. It owes much to the choice of local vision transformer architecture for both models. What is more interesting is that although UNETR is a second-best model in performance, its convergence speed is obviously the slowest, which is probably due to the direct usage of pure MSA resulting in the extra difficulty for model learning. The increasing number of parameters makes UNETR become unstable in the later stage and more prone to over-fitting.

\end{document}